\documentclass[11pt]{article}

\usepackage[numbers, compress]{natbib}

\usepackage[utf8]{inputenc} 
\usepackage[T1]{fontenc}    
\usepackage[left=3cm, right=3cm, top=4cm, bottom=4cm]{geometry}
\usepackage[pdftitle={Kernel methods through the roof: handling billions of points efficiently},
			pdfauthor={Giacomo Meanti}
			]{hyperref}
\usepackage{url}            
\usepackage{booktabs}       
\usepackage{amsfonts}       
\usepackage{nicefrac}       
\usepackage{microtype}      
\usepackage{amsmath}
\usepackage{mathtools}
\usepackage{graphicx}
\usepackage{bm}
\usepackage[inline]{enumitem}
\usepackage{etoolbox,siunitx}
\robustify\bfseries
\usepackage{multirow,makecell}
\usepackage{threeparttable}
\usepackage{caption}
\usepackage[labelformat=simple]{subcaption}

\usepackage{tikz}
\usetikzlibrary{matrix,shapes,shapes.geometric,decorations.pathreplacing,patterns,patterns.meta,fit,arrows.meta}

\pgfdeclarepattern{%
	name=hatch,
	parameters={\hatchsize,\hatchangle,\hatchlinewidth},
	bottom left={\pgfpoint{-.1pt}{-.1pt}},
	top right={\pgfpoint{\hatchsize+.1pt}{\hatchsize+.1pt}},
	tile size={\pgfpoint{\hatchsize}{\hatchsize}},
	tile transformation={\pgftransformrotate{\hatchangle}},
	code={
		\pgfsetlinewidth{\hatchlinewidth}
		\pgfpathmoveto{\pgfpoint{-.1pt}{-.1pt}}
		\pgfpathlineto{\pgfpoint{\hatchsize+.1pt}{\hatchsize+.1pt}}
		\pgfpathmoveto{\pgfpoint{-.1pt}{\hatchsize+.1pt}}
		\pgfpathlineto{\pgfpoint{\hatchsize+.1pt}{-.1pt}}
		\pgfusepath{stroke}
	}
}
\tikzset{%
	hatch size/.store in=\hatchsize,
	hatch angle/.store in=\hatchangle,
	hatch line width/.store in=\hatchlinewidth,
	hatch size=5pt,
	hatch angle=0pt,
	hatch line width=.5pt,
}

\usepackage{algorithm}
\usepackage{algpseudocode}
\usepackage{nicefrac}
\usepackage{adjustbox}
\usepackage{authblk}

\DeclarePairedDelimiterX{\norm}[1]{\lVert}{\rVert}{#1}

\DeclarePairedDelimiter\autobracket{(}{)}

\renewcommand{\O}[1]{\ensuremath{
		\mathcal{O}\autobracket*{ #1 }}}

\newcommand{\real}[1]{\mathbb{R}^{#1}}

\newcommand{\hilbert}{\ensuremath{\mathcal{H}}}

\DeclareMathOperator*{\argmin}{arg\,min}

\newcommand{\eye}{\ensuremath{\bm{I}}}
\newcommand{\chol}{\ensuremath{\mathrm{chol}}}

\title{
	Kernel methods through the roof: \\handling billions of points efficiently
}

\author[1]{Giacomo Meanti}
\author[1]{Luigi Carratino}
\author[1,2,3]{Lorenzo Rosasco}
\author[4]{Alessandro Rudi}
\affil[1]{\small \textit{MaLGa, DIBRIS, Università degli Studi di Genova, Genova, Italy}}
\affil[2]{\small \textit{Center for Brains, Minds and Machines, MIT, Cambridge, MA, USA}}
\affil[3]{\small \textit{Istituto Italiano di Tecnologia, Genova, Italy}}
\affil[4]{\small \textit{INRIA - Département d’Informatique de l’École Normale Supérieure - PSL Research University, Paris, France}}
\affil[ ]{ \small\texttt{giacomo.meanti@edu.unige.it}~~ \texttt{luigi.carratino@dibris.unige.it}~~ \texttt{lorenzo.rosasco@unige.it} \texttt{alessandro.rudi@inria.fr} }

\date{}

\begin{document}
\maketitle
\begin{abstract}
	
Kernel methods provide an elegant and principled approach to nonparametric learning, but so far could hardly be used in large scale problems, since na{\"i}ve implementations scale poorly with data size.
Recent advances have shown the benefits of a number of algorithmic ideas, for example combining optimization, numerical linear algebra and random projections.
Here, we push these efforts further to develop and test a solver that takes full advantage of GPU hardware.
Towards this end, we designed a preconditioned gradient solver for kernel methods exploiting both GPU acceleration and parallelization with multiple GPUs, implementing out-of-core variants of common linear algebra operations to guarantee optimal hardware utilization.
Further, we optimize the numerical precision of different operations and maximize efficiency of matrix-vector multiplications.
As a result we can experimentally show dramatic speedups on datasets with billions of points, while still guaranteeing state of the art performance.
Additionally, we make our software available as an easy to use library\footnote{\url{https://github.com/FalkonML/falkon}}.

%
\end{abstract}

\section{Introduction}

Kernel methods  provide non-linear/non-parametric extensions of many classical linear models in machine learning and statistics \cite{scholkopf02svm, cristianini}. The data are embedded 
via a non-linear map into a high dimensional feature space, so that  linear models in such a space effectively define non-linear models in the original space. 
This approach is appealing, since it naturally extends to models with infinitely many features, as long as the inner product 
in the feature space can be computed. 
In this case, the inner product is replaced by a positive definite kernel, and infinite dimensional models are reduced to finite dimensional problems.
The mathematics of kernel methods has its foundation in the rich theory of reproducing kernel Hilbert spaces \cite{schwartz64}, 
and the connection to linear models provides a gateway to deriving sharp statistical results \cite{steinwart09, smale01, kernelpca20, caponnetto07, bach13, sutherland18}. 
Further, kernel methods are tightly connected to Gaussian processes~\cite{gpbook}, and have recently being used to understand the properties of  deep learning models~\cite{ntk18, eigenpro17}.  It is not a surprise that kernel methods are among the most theoretically studied models.
From a numerical  point of view, they  reduce to  convex optimization problems that can be  solved with strong guarantees. 
The corresponding algorithms provide excellent results on a variety of data-sets, but most implementations
are limited to problems of small/medium size, see discussion in~\cite{steinwart2008support}, Chapter~11. Most methods require handling a kernel matrix  
quadratic in the sample size. Hence, dealing with datasets of size $10^4$ to $10^5$ is challenging, 
while larger datasets are typically out of reach. A number of approaches have been considered to alleviate these computational bottlenecks. 
Among others, random features~\cite{rahimi08,rahimi09,yang12,le13,dai14,cutajar17} and the Nystr\"om method are often used~\cite{williams01, smola00}, see also~\cite{drineas05, kumar12, gittens16, fasterkrr17, zhang15, hierachical17}.
While different, both these approaches consider random projections to reduce the problem size and hence computational costs. 
Renewed interest in approximate kernel methods was also spurred by recent theoretical results proving that computational 
gains can possibly be  achieved with no loss of accuracy, see e.g.~\cite{unifiedrf19, intheory18, rudirosascorf, bach13, lessismore15, danielekmeans18}.%

{%
	\begin{figure}[t]
		\centering
		\includegraphics[width=\textwidth]{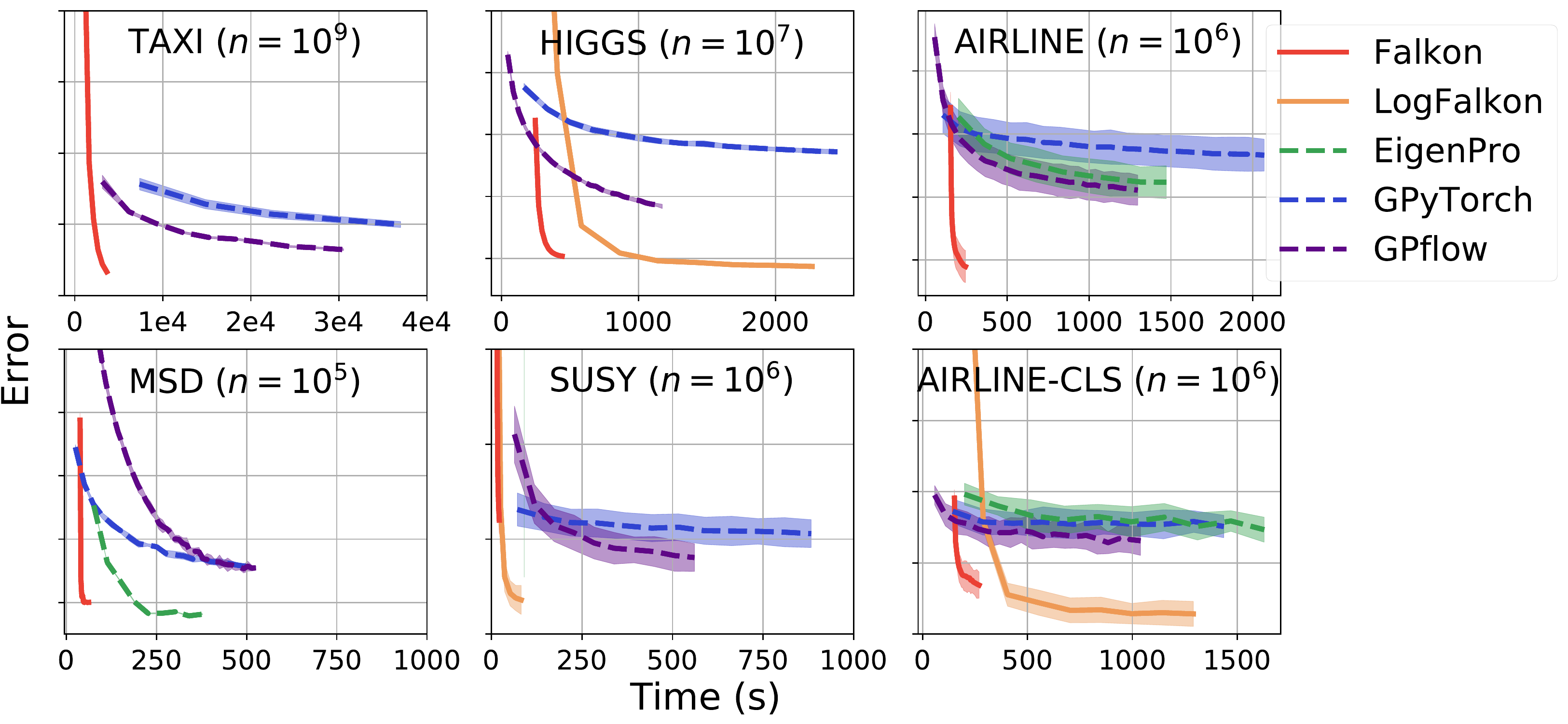}
		\caption{Benchmarks of kernel solvers on large scale datasets with millions and billions points (see Section~\ref{sec:exp}). Our approach (red and yellow lines) consistently achieves state of the art accuracy in minutes.}
		\label{fig:results}
	\end{figure}%
}%

In this paper, we investigate the practical consequences of this line of work, developing and testing large scale  kernel methods that can run efficiently on billions of points. Following~\cite{rudi2017falkon} we use a Nystr\"om approach to reduce the problem size and also to derive a preconditioned gradient solver for kernel methods. 
Indeed, we focus on smooth loss functions where such approaches are natural.
Making these algorithmic ideas practical and capable of exploiting the GPU, requires developing a number of computational solutions, 
borrowing ideas not only from optimization and numerical analysis but also from scientific and high performance computing \cite{ltaief10magma,krylovhpc15,svmgpu08}. 
Indeed, we design preconditioned conjugate gradient solvers that take full advantage of both GPU acceleration and parallelization with multiple GPUs, implementing out-of-core variants of common linear algebra operations to guarantee optimal hardware utilization.
We further optimize the numerical precision of different operations and investigate ways to perform matrix-vector multiplications most efficiently. 
The corresponding implementation is then tested extensively on a number of datasets ranging from millions to billions of points.
For comparison, we focused on other available large scale kernel implementations that do not require data splitting, or multiple machines.
In particular, we consider Eigenpro~\cite{eigenpro2} which is an approach similar to the one we propose, and GPyTorch~\cite{gpytorch18} and GPflow~\cite{GPflow2020} which come from the Gaussian process literature. While these latter solutions allow also for uncertainty quantification, we limit the comparison to prediction.
We perform a systematic empirical evaluation running an extensive series of tests.
Empirical results show that  indeed our  approach can process  huge datasets in minutes  and obtain state of the art performances,  comparing favorably to other solutions, both in terms of efficiency and accuracy. More broadly, these results confirm and extend the observations made in \cite{eigenpro17, eigenpro2}, that kernel methods can now be seamlessly and effectively deployed on large scale problems. To make these new solutions readily available,
the corresponding code is distributed as an easy to use library developed on top of PyTorch~\cite{pytorch}.\\
The rest of the paper is organized as follows. In Section~\ref{sec:background}, we provide some background on the considered approaches. In Section~\ref{sec:methods}, we detail the main algorithmic solutions in our implementation, whereas the last section is devoted to assessing  the practical advantages.

\section{Background}\label{sec:background}
Supervised learning is the problem of inferring an input-output function, given finitely many input-output pairs. In statistical learning theory the data $(x_i, y_i)_{i=1}^{n}$ are assumed to be sampled independently from a probability distribution $\rho$, and a loss function $\ell(y, f(x))$ is fixed measuring the cost of predicting $f(x)$ in place of $y$.
The examples we consider are the squared $(y-f(x))^2$ and the logistic loss $\log(1+e^{-yf(x)})$.
Then, a good function $f$ should minimize the expected loss
\begin{equation}\label{eq:learning-problem}
L(f) = \int \ell \big(f(x), y\big) d\rho(x, y).
\end{equation}
A basic approach to solve the problem is empirical risk minimization, based on the idea of replacing the above expectation with an empirical average.
Further, the search of a solution needs to be restricted to a suitable space of hypothesis, a simple example being linear functions $f(x)=w^\top x$. 
Kernel methods extend this idea by considering a non linear feature map $x\mapsto \Phi(x)\in {\cal F}$ and functions of the form $f(x)=w^\top \Phi(x)$.
Here $\Phi(x)\in {\cal F}$ can be seen as a feature representation in some space of features. 
The function space $\hilbert$ thus defined is called reproducing kernel Hilbert space~\cite{scholkopf01}. 
If we denote by $\norm{f}_\hilbert$ its norm then regularized empirical risk minimization is given by 
\begin{equation}\label{eq:flambda_estimator}
\hat{f}_\lambda = \argmin_{f\in\hilbert}\dfrac{1}{n} \sum_{i=1}^n \ell\big(f(x_i), y_i\big) + \lambda \norm{f}^2_\hilbert,
\end{equation}
where the penalty term $\norm{f}_\hilbert$ is meant to prevent possible instabilities and $\lambda \geq 0$ is a hyperparameter.
From a statistical point of view the properties of the estimator $\hat{f}_\lambda$ are well studied, see e.g.~\cite{steinwart09, caponnetto07, shwartz-bendavid}. 
Under basic assumptions, for  $\lambda = \O{1/\sqrt{n}}$,  it holds with high probability that
\begin{equation}\label{eq:bound} 
L(\hat{f}_\lambda) - \inf_{f \in \hilbert} L(f) = \O{n^{-1/2}}.
\end{equation} 
This bound is sharp, but can be improved under further assumptions \cite{caponnetto07, steinwart09}. Here, we use it for reference.
From a computational point of view, the key fact is that it is possible to compute a solution also if $\Phi(x)$ is an infinite feature vector, as long as 
the kernel $k(x,x')=\Phi(x)^\top \Phi(x')$ can be computed~\cite{scholkopf02svm}. The Gaussian kernel $\exp(-\norm{x-x'}^2/2\sigma^2)$ is a basic example. Indeed, by the representer theorem~\cite{representer70,scholkopf01},
$
\hat f _\lambda(x) = \sum_{i=1}^n\alpha_i k(x, x_i),
$
so Problem~\eqref{eq:flambda_estimator} can be replaced with a finite dimensional problem on the coefficients. 
Its solution depends on the considered loss, but typically involves handling the kernel matrix 
$K_{nn}\in \real{n\times n}$ with entries $k(x_i, x_j)$, which becomes prohibitive as soon as $n\sim 10^5$ (although multi-GPU approaches~\cite{wang19million} have been recently shown to scale to $10^6$ points).
In the following, we focus on  Nystr\"om approximation,  considering  functions of the form
\begin{equation}\label{eq:nystrom-solution}
f(x) =  \sum_{i=1}^m \alpha_i k(x, \tilde{x}_i),
\end{equation}
where  $\{\tilde{x}_1, \dots, \tilde{x}_m\} \subset \{x_1, \dots, x_n\}$ are  inducing points  sampled uniformly at random.
As we discuss next, this approach immediately yields computational gains. Moreover, recent theoretical results show that 
the basic bound in~\eqref{eq:bound} still holds taking 
as few as $m=\O{\sqrt{n}}$ inducing points~\cite{lessismore15, mareau19}. With these observations in mind, we next 
illustrate how these algorithmic ideas can be developed considering first  the square loss and than the logistic loss.\\%
\begin{algorithm}[t]
	\caption{Pseudocode for the Falkon algorithm. \label{alg:falkon}}
	\begin{minipage}{0.53\linewidth}{
			\footnotesize
			\begin{algorithmic}[1]
				\footnotesize
				\Function{Falkon}{$X \in \real{n\times d}, \bm{y} \in \real{n}, \lambda, m, t$}
				\State $X_{m} \gets$ \Call{RandomSubsample}{$X, m$} 
				\State $T, A \gets $ \Call{Preconditioner}{$X_m, \lambda$}
				\Function{LinOp}{$\bm{\beta}$} 
				\State $\bm{v} \gets A^{-1}\bm{\beta}$
				\State $\bm{c} \gets k(X_m, X)k(X, X_m)T^{-1}\bm{v}$
				\State \textbf{return} $A^{-\top} T^{-\top} \bm{c} + \lambda n \bm{v}$
				\EndFunction
				
				\State $R \gets A^{-\top}T^{-\top}k(X, X_m)\bm{y}$ 
				\State $\bm{\beta} \gets $ \Call{ConjugateGradient}{$\textsc{LinOp}, R,t$} 
				\State \textbf{return} $T^{-1}A^{-1} \bm{\beta}$ 
				\EndFunction
			\end{algorithmic}
		}
	\end{minipage}~~%
	\begin{minipage}{0.46\linewidth}
		\footnotesize
		\begin{algorithmic}[1]
			\setcounter{ALG@line}{12}
			\footnotesize
			\Function{Preconditioner}{$X_m \in \real{m\times d}, \lambda$}
			\State $K_{mm} \gets k(X_m, X_m)$ \label{line:kernel} 
			\State $T \gets \chol(K_{mm})$ \label{line:chol1} 
			\State $K_{mm} \gets \nicefrac{1}{m} T T^\top  + \lambda \eye$ \label{line:lauum_bg} 
			\State $A \gets \chol(K_{mm})$ \label{line:chol2}
			\State \textbf{return} $T, A$ 
			\EndFunction
		\end{algorithmic}
		
		$ $
		
		Note: LinOp performs the multiplication $\tilde{P}^\top H \tilde{P} \beta$ as in Eq.~\eqref{eq:cg-iter-non-opt}, via matrix-vector products.
		
		$ $
	\end{minipage}%
\end{algorithm}%
{\bf Squared loss.}
This choice corresponds to kernel ridge regression (KRR). Since both the loss and penalty are quadratic, solving KRR reduces to solving a linear system. 
In particular, letting $\bm{y}=(y_1, \dots, y_n)$, we obtain 
$
(K_{nn}+ \lambda n I)\bm{\alpha} = \bm{y},
%
$
for the coefficients $\bm{\alpha}=(\alpha_1, \dots, \alpha_n)\in\real{n}$ in the solution of the problem in Eq.~\eqref{eq:flambda_estimator}, while using the Nystr\"om approximation \eqref{eq:nystrom-solution} we get
\begin{equation}\label{eq:nkrr-system}
(K_{nm}^\top K_{nm} + \lambda n K_{mm})\bm{\alpha} = K_{nm}^\top \bm{y}, 
\end{equation}
for  $\bm{\alpha}=(\alpha_1, \dots, \alpha_m)\in\real{m}$.
The first linear system can be solved directly in $\O{n^3}$ time and $\O{n^2}$ space. In turn, Eq.~\eqref{eq:nkrr-system} can be solved directly in $\O{nm^2 + m^3}$ time and $\O{m^2}$ space (if the $K_{nm}$ matrix is computed in blocks).
It is well known, that for large linear systems iterative solvers are preferable \cite{saad2003iterative}. 
Further, the convergence of the latter can be greatly improved by considering preconditioning.
The na{\"i}ve preconditioner $P$ for problem~\eqref{eq:nkrr-system} is such that $PP^\top  = (K_{nm}^\top K_{nm} + \lambda n K_{mm})^{-1}$, and as costly to compute as the original problem. Following~\cite{rudi2017falkon} it can be approximated using once again the Nystr{\"o}m method to obtain 
\begin{equation}\label{eq:flk_preconditioner}
\tilde{P}\tilde{P}^\top  = ( \tfrac{n}{m} K^2_{mm} + \lambda n K_{mm} )^{-1}
\end{equation}
since $K^2_{mm} \approx K_{nm}^\top K_{nm}$. Next, we follow again~\cite{rudi2017falkon} and combine the above preconditioning with conjugate gradient (CG).
The pseudocode of the full procedure is given in Algorithm~\ref{alg:falkon}. Indeed, as shown in~\cite{rudi2017falkon} $\O{\log{n}}$ CG steps are sufficient to achieve the bound in~\eqref{eq:bound}. Then with this approach, the total computational cost to achieve optimal statistical bounds is  $\O{n\sqrt{n}\log n}$ in time, and in $\O{n}$ in memory, making it ideal for large scale scenarios. 
The bulk of our paper is devoted to developing solutions to efficiently implement and deploy Algorithm~\ref{alg:falkon}.\\ 
%
%
{\bf Logistic loss.}  The above ideas extend to the logistic loss and more generally to self-concordant loss functions, including the softmax loss~\cite{marteau2019beyond}. For reasons of space, we detail this case in Appendix~\ref{app:logistic} and sketch here the main ideas.
In this case, iterative solvers are the default option since there is no closed form solution. Nystr\"om method can be used a first time to reduce the size of the problem, and then a second time to derive an approximate Newton step~\cite{mareau19}. 
More precisely,  at every step preconditioned conjugate gradient descent is run for a limited number of iterations with a decreasing value of $\lambda$, down to the desired regularization level.
In practice, this requires running Algorithm~\ref{alg:falkon} multiple times with small number of iterations $t$ and with decreasing $\lambda$.
%
Making these ideas practical requires efficiently implementing and deploying Algoritm~\ref{alg:falkon}, making full use of the available computational architectures.
This the core of our contribution that we detail in the next section.

\section{Reformulating kernel solvers for multi-core/multi-GPU architectures}\label{sec:methods}

\begin{figure}[b]
	\centering
	\begin{minipage}{0.4\textwidth}
		\centering
		\includegraphics[width=0.8\textwidth]{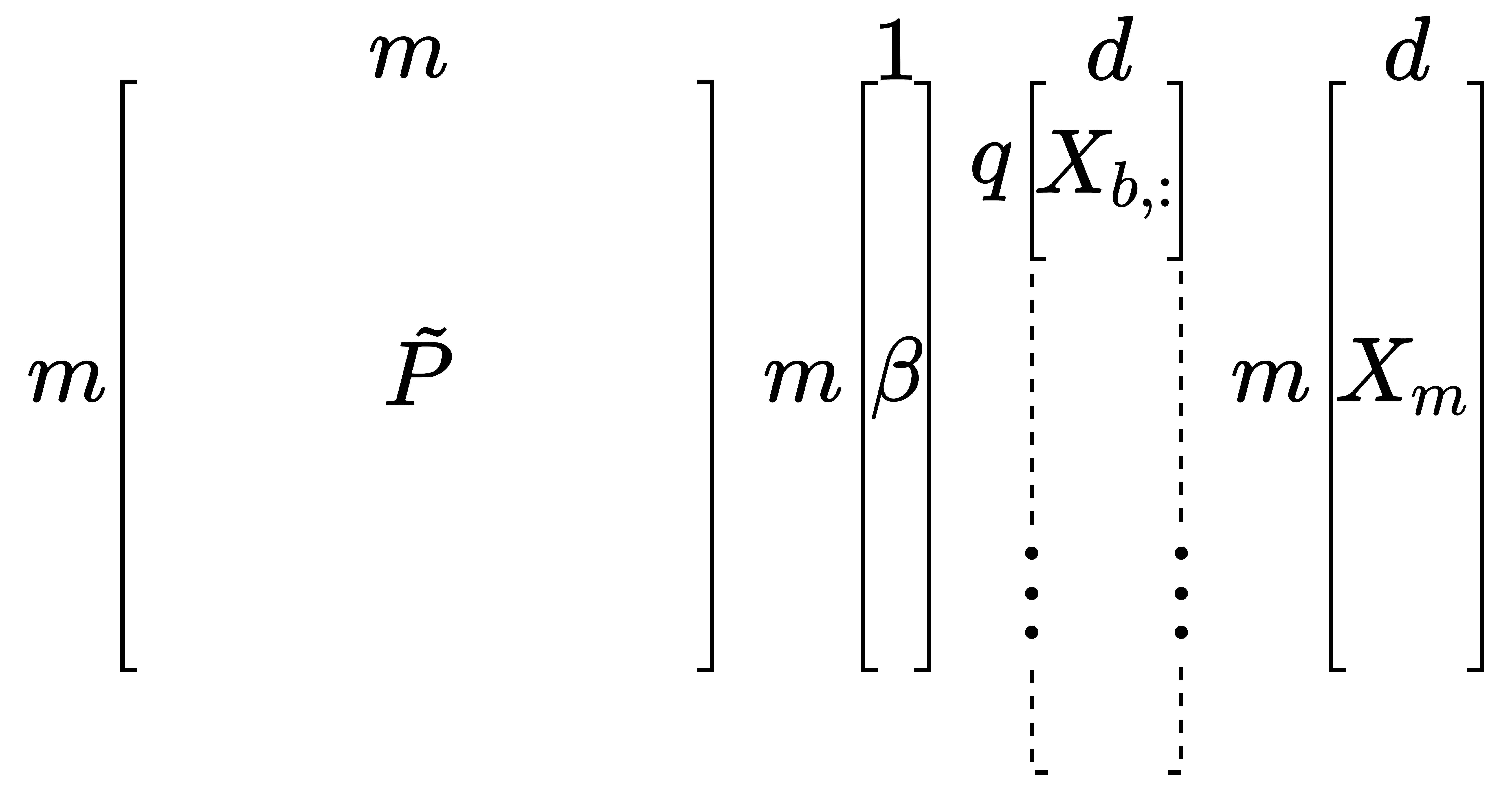}
		\caption{Structure of RAM allocation.}
		\label{fig:mem-buffers}
	\end{minipage}%
	\hfill
	\begin{minipage}{0.58\textwidth}
		\centering
		\begin{tikzpicture}[%
	line width=0.5pt,
	mstyle/.style={%
		text height=1.3ex,
		text depth=0.45ex,
		text width=5.0ex,
		align=center,
		column sep=0pt,
		row sep=0pt,
		nodes in empty cells,
	}
]

\matrix[matrix of nodes, mstyle] at (0, 0) (M) {%
	CPU$\rightarrow$GPU & & & & & &\\
	Compute         & & & & & &\\
	GPU$\rightarrow$CPU & & & & & &\\
	                & & & & & &\\
};

\fill [red!50] (M-1-3.south west) rectangle (M-1-3.north east);
\fill [red!50] (M-2-4.south west) rectangle (M-2-4.north east);
\fill [red!50] (M-3-5.south west) rectangle (M-3-5.north east);

\fill [green!50] (M-1-4.south west) rectangle (M-1-4.north east);
\fill [green!50] (M-2-5.south west) rectangle (M-2-5.north east);
\fill [green!50] (M-3-6.south west) rectangle (M-3-6.north east);

\fill [blue!50] (M-1-5.south west) rectangle (M-1-5.north east);
\fill [blue!50] (M-2-6.south west) rectangle (M-2-6.north east);
\fill [blue!50] (M-3-7.south west) rectangle (M-3-7.north east);

\draw[line width=1pt] (M-1-3.north west) -- (M-3-3.south west);
\draw[line width=1pt,dashed] (M-1-4.north west) -- (M-3-4.south west);
\draw[line width=1pt,dashed] (M-1-5.north west) -- (M-3-5.south west);
\draw[line width=1pt,dashed] (M-1-6.north west) -- (M-3-6.south west);
\draw[line width=1pt,dashed] (M-1-7.north west) -- (M-3-7.south west);
\draw[line width=1pt] (M-1-7.north east) -- (M-3-7.south east);

\draw[-latex] (M-4-4.center) to (M-4-6.center);
\draw (M-4-5)++(0pt,-8pt) node {Time};

\end{tikzpicture}
		\caption{Overlapping memory transfers and computation.}
		\label{fig:super-scalar}
	\end{minipage}%
\end{figure}%

GPU machines have a peculiar architecture with rather different properties than the standard von Neumann computer, in particular they are characterized by highly parallel computational power, relatively small local accelerator memory and slow memory transfer to/from the accelerator compared to their computational speed~\cite{wilt2013cuda}.
In their standard definition, kernel methods require large amounts of memory with a low density of operations per byte of memory used. 
This opens the question of how to adapt methods with low operation density to platforms designed to be extremely efficient with very high density of operations per byte.
With this in mind, we started considering the state of the art kernel solver with minimal computational requirements for optimal guarantees (described at a high level in Algorithm~\ref{alg:falkon}), with the goal to reformulate its computational structure to dramatically increase the density of operations per byte, and reduce as much as possible the required memory use / transfers.
To achieve this goal, we use a number of carefully designed computational solutions which systematically reduce the impact of the inherent bottlenecks of multi-core/multi-GPU architectures, while leveraging their intrinsic potential. In particular in the rest of this section we will focus on \begin{enumerate*}[label=(\alph*)]
	\item minimizing the memory footprint of the solver, which has long been the main bottleneck for kernel methods, and is the main limitation encountered by current kernel solvers,
	\item dealing with limited memory on the GPU,
	\item reaching the highest possible accelerator utilization, parallelizing memory transfers and computation,
	\item using the enhanced capabilities of GPUs with reduced-precision floating point data.
\end{enumerate*}

\subsection{Overcoming RAM memory bottleneck}\label{sec:RAM-memory}
\begin{figure}
	\centering
	\resizebox{!}{3.5cm}{\begin{tikzpicture}[%
	line width=0.5pt,
	mstyle/.style={%
		text height=2.5ex, 
		text depth=0.6ex,
		text width=3ex, 
		align=center,
		column sep=0pt,
		row sep=0pt,
		left delimiter=[,
		right delimiter=],
		nodes in empty cells,
	},
	farrow/.style={line width=1pt, arrows={-{Latex[length=3mm]}}},
	triangle/.style={black!15}
]

\matrix[matrix of nodes, mstyle] at (0,0) (M1){%
	& &\\
	& &\\
    & &\\
};
\fill[triangle] (M1-1-1.north west) -- (M1-1-3.north east) -- (M1-3-3.south east) -- cycle;
\node at (M1-2-2.center) {$K_{mm}$};

\matrix[matrix of nodes, mstyle] at (3.5,0) (M2){%
	& &\\
	& &\\
	& &\\
};
\fill[triangle] (M2-1-1.north west) -- (M2-1-3.north east) -- (M2-3-3.south east) -- cycle;
\node at (M2-2-2.north east) {$T$};
\node[yshift=-5pt, xshift=0pt] at (M2-3-2.north west) {$\mathrm{tril}(K_{mm})$};

\matrix[matrix of nodes, mstyle] at (7,0) (M3){%
	& &\\
	& &\\
	& &\\
};
\fill[triangle] (M3-1-1.north west) -- (M3-1-3.north east) -- (M3-3-3.south east) -- cycle;
\node at (M3-2-2.north east) {$T$};
\node[yshift=-8pt, xshift=0pt] at (M3-3-2.north west) {$\frac{1}{m}TT^T + \lambda \eye$};

\matrix[matrix of nodes, mstyle] at (10.5,0) (M4){%
	& &\\
	& &\\
	& &\\
};
\fill[triangle] (M4-1-1.north west) -- (M4-1-3.north east) -- (M4-3-3.south east) -- cycle;
\node at (M4-2-2.north east) {$T$};
\node[yshift=0pt, xshift=-5pt] at (M4-2-1.south east) {$A^T$};

\draw[farrow] (M1-1-2.north) to[bend left] node[midway,above,inner sep=0pt] {Cholesky} (M2-1-2.north);
\draw[farrow] (M2-1-2.north) to[bend left] node[midway,above,inner sep=0pt] {Matrix multiply} (M3-1-2.north);
\draw[farrow] (M3-1-2.north) to[bend left] node[midway,above,inner sep=0pt] {Cholesky} (M4-1-2.north);

\end{tikzpicture}}
	\caption{Evolution of the preconditioner matrix in memory.}
	\label{fig:preconditioner}
\end{figure}
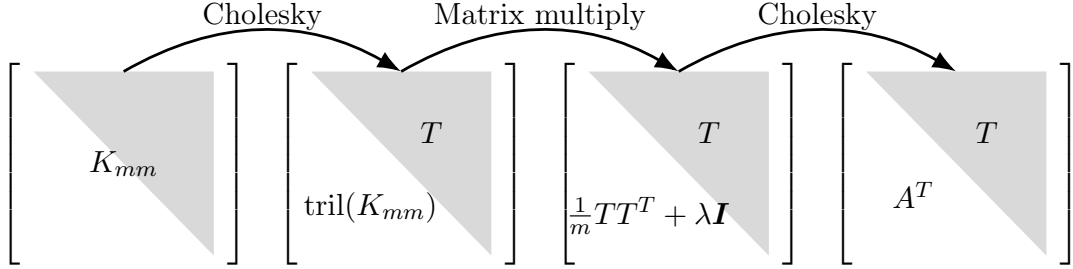

Kernel solvers that use the Nystr{\"o}m method need the matrices $K_{mm}$ and $K_{nm}$. Since $K_{nm}$ is used only in matrix-vector products, we can avoid constructing it explicitly (as we shall see in the following paragraphs) which leaves us to deal with the $K_{mm}$ matrix. When $m$ is large, it is crucial to carefully manage the  memory needed for this task: 
in our implementation we only ever allocate one $m\times m$ matrix, and overwrite it in different steps to calculate the preconditioner.
Indeed, choosing an appropriate form of the preconditioner, the matrix $K_{mm}$ itself is not needed in the conjugate gradient iteration. 
Figure~\ref{fig:mem-buffers} shows the total memory usage, which consists of the preconditioner occupying approximately $90\%$ of the memory (see last paragraph of Sect.~\ref{sec:RAM-memory}), the weight vector $\bm{\beta}$ and two buffers holding (part of) the $m$ inducing points and a data batch needed to compute $K_{nm}$.\\
%
\noindent{\bf In-place computation and storage of the preconditioner.}
The preconditioner $\tilde{P}$ of Eq.~\eqref{eq:flk_preconditioner} is used to solve a linear system of the form $\tilde{P}^\top  H \tilde{P}\bm{\beta} = \tilde{P}^\top K_{mn}\bm{y}$ with $H = K_{mn}K_{nm} + \lambda n K_{mm}$ and $\bm{\beta} = \tilde{P}^{-1}\bm{\alpha}$.
$\tilde{P}$ can be decomposed into two triangular matrices obtained via Cholesky decomposition of $K_{mm}$,
\begin{equation}\label{eq:flk-prec-chol}
\tilde{P} = \tfrac{1}{\sqrt{n}}T^{-1}A^{-1},\qquad T = \chol(K_{mm}), \qquad A = \chol(\tfrac{1}{m} TT^\top  + \lambda \eye_m).
\end{equation}
All operations are performed in-place allocating a single $m\times m$ matrix as shown in Figure~\ref{fig:preconditioner} and as described next:
\begin{enumerate*}[label=(\alph*)]
	\item a matrix of dimension $m\times m$ is allocated in memory;
	\item the $K_{mm}$ kernel is computed in blocks on the GPU and copied to the matrix;
	\item {\em in-place} Cholesky decomposition of the upper triangle of $K_{mm}$ is performed on the GPU (if the kernel does not fit GPU memory an out-of-core algorithm is used, see later sections);
	\item the product $TT^\top$  is computed in blocks via GPU and stored in the lower part;
	\item out-of-core in-place Cholesky decomposition is performed on the lower triangle to get $A^\top$.
\end{enumerate*}
Additional care is needed to take into account the matrix diagonal, not described here for brevity.\\
\noindent{\bf Elimination of the storage of $K_{mm}$.} Considering more carefully the matrix $\tilde{P} (K_{nm}^\top K_{nm} + \lambda n K_{mm}) \tilde{P}$ with $\tilde{P}$ as in Eq.~\eqref{eq:flk-prec-chol}, we observe that the occurrences of $K_{mm}$ cancel out. Indeed $(T^{-1})^{\top}K_{mm}T^{-1} = \eye$ since $K_{mm}=T^\top  T$ by Eq.~\ref{eq:flk-prec-chol}. Then, the following characterization allows to overwrite $K_{mm}$ when calculating the preconditioner.
\begin{align}
\tilde{P}^\top H\tilde{P}\beta &=~~~ (A^{-1})^{\top} (T^{-1})^{\top} (K_{nm}^\top K_{nm} + \lambda n K_{mm})T^{-1}A^{-1}\bm{\beta} \label{eq:cg-iter-non-opt}\\
& =~~~ (A^{-1})^{\top} [ (T^{-1})^{\top}  K_{nm}^\top  K_{nm} T^{-1}+ \lambda n I ]A^{-1}\bm{\beta}. \label{eq:cg-iter}
\end{align}\\
%
\noindent{\bf Blockwise $K_{nm}$-vector product on GPU.}
The conjugate gradient algorithm will repeatedly execute Eq.~\eqref{eq:cg-iter} for different $\bm{\beta}$. The most expensive operations are the matrix-vector products $K_{nm}^\top (K_{nm} \bm{v})$ for an arbitrary vector $\bm{v} \in \real{m\times 1}$ which -- if computed explicitly -- would require $n \times m$ memory. 
However, it is possible to split the input data $X \in \real{n\times d}$ in $B$ batches of $q$ rows each $\{X_{b,:} \in \real{q \times d}\}_{b=1}^B$, so that matrix-vector products can be accumulated between batches using the formula $\sum_{b=1}^B k(X_{b,:}, X_m)^\top  (k(X_{b,:}, X_m) \bm{v})$.
The matrix blocks to be held in memory are summarized in Figure~\ref{fig:mem-buffers} for a total size of $m \times (m + d + 1) + q\times d$ where $q$ can be small under memory pressure, or large for greater performance.
It is important to note that $k(X_{b,:}, X_m)$ is never stored in main memory, as all operations on it are done on the GPU.

\subsection{Fitting in GPU memory and dealing with multiple GPUs}
While the main RAM might be a bottleneck, GPUs have an even smaller amount of memory, and  another level of splitting is needed to exploit their speed. For example, a typical architecture has 256GB of RAM and 4 GPUs with 16GB ram each; a preconditioner with $m=2 \times 10^5$ occupies \SI{150}{\giga\byte} and $K_{nm}$ with $n=10^7$ would need \SI{2000}{\giga\byte} of memory if stored. 
So we need to deal with both efficient computation of $K_{nm}$-vector product in chunks that fit a GPU, and with the computation of the preconditioner that usually does not fit in GPU memory. Operations based on a large storage layer (main RAM) and a small but fast layer (GPU) are called out-of-core (OOC) operations. 
However, common machine learning libraries such as Tensorflow~\cite{tensorflow} or PyTorch~\cite{pytorch} do not implement OOC versions of the required matrix operations, leaving potentially complex implementations to the users. 
Hence, in our library, we provide these implementations in easily reusable form.
It is important to note that splitting our workload to fit in GPU also provides an easy path to parallelization in a multi-GPU system: new chunks of computation are assigned to the first free GPU, effectively redistributing the workload between multiple accelerators when available. \\
\noindent{\bf Optimized block decomposition for out-of-core $K_{nm}$-vector multiplication.} 
As seen in the previous section, matrix-vector products can be split along the dimension $n$, resulting in independent chunks of work that need to be summed up at the end. The OOC product between a kernel matrix and a vector proceeds by: \begin{enumerate*}[label=(\alph*)] \item transferring a block of data onto the device, \item computing the kernel on device and multiplying it by the vector, \item copying the result back to the host.\end{enumerate*} This sequence of operations minimizes expensive data-transfers between host and device since the kernel matrix is never moved.
In particular, the computation is also split along dimensions $m$ and $d$, to maximize the ratio between computational complexity and transfer time: i.e., maximizing $\frac{qrs}{qs + ds}$ subject to $qs + ds \leq G$, where $q$, $r$ and $s$ are the batch dimensions along $n$, $m$ and $d$ respectively, and $G$ is the available GPU memory.\\%
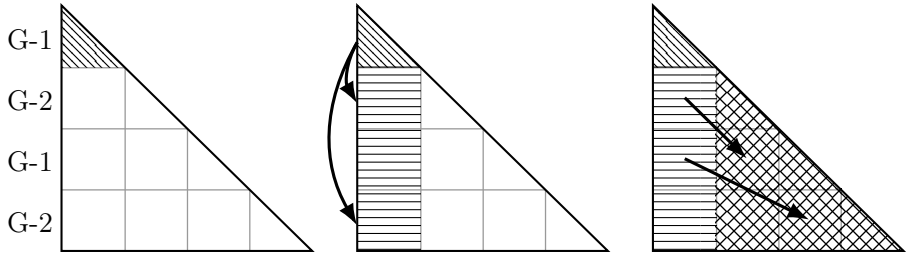
\begin{figure}[t]
	\centering
	\resizebox{!}{3.5cm}{
		\begin{tikzpicture}[%
	line width=0.5pt,
	mstyle/.style={%
		text height=2.7ex, 
		text depth=0.75ex, 
		text width=3.55ex, 
		align=center,
		column sep=0pt,
		row sep=0pt,
		nodes in empty cells,
	},
	farrow/.style={line width=1.2pt, arrows={-{Latex}}},
]

\matrix[matrix of nodes, mstyle, align=left, text width=4ex] at (-0.1,0) (MT){%
	G-1 \\
	G-2 \\
	G-1 \\
	G-2 \\
};

\matrix[matrix of nodes, mstyle] at (2,0) (M){%
	& & & \\
	& & & \\
	& & & \\
	& & & \\
};

\draw[black!40] (M-1-1.south east) -- (M-4-1.south east);
\draw[black!40] (M-2-2.south east) -- (M-4-2.south east);
\draw[black!40] (M-3-3.south east) -- (M-4-3.south east);
\draw[black!40] (M-1-1.south west) -- (M-1-1.south east);
\draw[black!40] (M-2-1.south west) -- (M-2-2.south east);
\draw[black!40] (M-3-1.south west) -- (M-3-3.south east);

\draw[thick] (M-1-1.north west) -- (M-4-1.south west) -- (M-4-4.south east) -- cycle;

\fill[pattern=north west lines] (M-1-1.north west) -- (M-1-1.south west) -- (M-1-1.south east) -- cycle;

\matrix[matrix of nodes, mstyle] at (6,0) (M2){
	& & & \\
	& & & \\
	& & & \\
	& & & \\
};

\draw[black!40] (M2-1-1.south east) -- (M2-4-1.south east);  
\draw[black!40] (M2-2-2.south east) -- (M2-4-2.south east);
\draw[black!40] (M2-3-3.south east) -- (M2-4-3.south east);
\draw[black!40] (M2-1-1.south west) -- (M2-1-1.south east);  
\draw[black!40] (M2-2-1.south west) -- (M2-2-2.south east);
\draw[black!40] (M2-3-1.south west) -- (M2-3-3.south east);

\draw[thick] (M2-1-1.north west) -- (M2-4-1.south west) -- (M2-4-4.south east) -- cycle;

\fill[pattern=north west lines] (M2-1-1.north west) -- (M2-1-1.south west) -- (M2-1-1.south east) -- cycle;
\foreach \x in {2,...,4}
{
	\fill[pattern=horizontal lines] (M2-\x-1.north west) -- (M2-\x-1.north east) -- (M2-\x-1.south east) -- (M2-\x-1.south west) -- cycle;
	
}
\draw[farrow] (M2-1-1.mid west) to[bend right] (M2-2-1.mid west);
\draw[farrow] (M2-1-1.mid west) to[bend right] (M2-4-1.mid west);

\matrix[matrix of nodes, mstyle] at (10,0) (M3){
	& & & \\
	& & & \\
	& & & \\
	& & & \\
};

\draw[black!40] (M3-1-1.south east) -- (M3-4-1.south east);  
\draw[black!40] (M3-2-2.south east) -- (M3-4-2.south east);
\draw[black!40] (M3-3-3.south east) -- (M3-4-3.south east);
\draw[black!40] (M3-1-1.south west) -- (M3-1-1.south east);  
\draw[black!40] (M3-2-1.south west) -- (M3-2-2.south east);
\draw[black!40] (M3-3-1.south west) -- (M3-3-3.south east);

\draw[thick] (M3-1-1.north west) -- (M3-4-1.south west) -- (M3-4-4.south east) -- cycle;

\fill[pattern=north west lines] (M3-1-1.north west) -- (M3-1-1.south west) -- (M3-1-1.south east) -- cycle;
\foreach \x in {2,...,4}
{
	\fill[pattern=horizontal lines] (M3-\x-1.north west) -- (M3-\x-1.north east) -- (M3-\x-1.south east) -- (M3-\x-1.south west) -- cycle;
	
}
\foreach \x in {2,...,4}
{
	\fill[pattern={hatch}] (M3-\x-\x.north west) -- (M3-\x-\x.south west) -- (M3-\x-\x.south east) -- cycle;
}
\fill[pattern=hatch] (M3-3-2.north west) -- (M3-3-2.north east) -- (M3-3-2.south east) -- (M3-3-2.south west) -- cycle;
\fill[pattern=hatch] (M3-4-2.north west) -- (M3-4-2.north east) -- (M3-4-2.south east) -- (M3-4-2.south west) -- cycle;
\fill[pattern=hatch] (M3-4-3.north west) -- (M3-4-3.north east) -- (M3-4-3.south east) -- (M3-4-3.south west) -- cycle;
\draw[farrow] (M3-2-1.center) to (M3-3-2.center);
\draw[farrow] (M3-3-1.center) to (M3-4-3.center);

\end{tikzpicture}
	}
	\caption{Three phases of the block Cholesky decomposition for updating the first column. Arrows indicate inter-GPU memory transfers between accelerators G-1 and G-2.}
	\label{fig:potrf}
\end{figure}%
\noindent{\bf Out-of-core multi-GPU Cholesky decomposition.}
Other operations, such as Cholesky decomposition and triangular matrix multiplication (lines \ref{line:chol1},~\ref{line:lauum_bg},~\ref{line:chol2} of Algorithm~\ref{alg:falkon}), can also benefit from GPU execution. 
Here we describe, at a high level, our algorithm for multi-GPU OOC Cholesky decomposition inspired by~\cite{ltaief10magma, wu18parallel}. We leave further details to Appendix~\ref{app:ooc}.
Consider a symmetric matrix $A$, split into $B\times B$ tiles $A_{ij} \in \real{t\times t}, i \in [B], j \in [B]$, assumed of equal size for brevity. We want a factorization $A = L L^\top$, where $L$ is lower triangular, with the formula $A_{i,j} = \sum_{k=1}^j L_{i,k}L_{j,k}^\top$.
The algorithm runs in-place, updating one column of $A$ at a time. Each column update proceeds in three steps, illustrated in Figure~\ref{fig:potrf}. Clearly $A_{1,1} = L_{1,1}L_{1,1}^\top $ so we compute $L_{1,1}$ by a Cholesky decomposition on tile $A_{1,1}$ which is small and can be done entirely on the GPU (e.g. with cuSOLVER~\cite{cusolver}).
Then we consider the other tiles of the first block column of $L$ for which $A_{j,1} = L_{j,1}L_{1,1}^\top $ with $j > 1$. Since we know $L_{1,1}$ from the first step, we obtain $L_{j,1} = A_{j,1}L_{1,1}^{-\top} $ for all $j > 1$ by solving a triangular system (on the GPU). 
Finally the first block column of $L$ is used to update the trailing submatrix of $A$. Note that $A_{i,j} = \sum_{k=1}^j L_{i,k}L_{j,k}^\top  = L_{i,1}L_{j,1}^\top  + \sum_{k=2}^jL_{i,k}L_{j,k}^\top $ for $2 \le j \le i$, so we can update the trailing submatrix as $A_{i,j} = A_{i,j} - L_{i,1}L_{j,1}^\top $. 
We implemented a parallel version of the above algorithm which distributes block-rows between the available processors in a 1D block-cyclic way (e.g.~Figure~\ref{fig:potrf} (left): rows 1 and 3 are assigned to GPU-1, rows 2 and 4 are assigned to GPU-2). 
For each column update, one processor executes the first step and transfers the result to the others (the arrows in Figure~\ref{fig:potrf}), which can then execute step 2 in parallel. To update the trailing matrix, further data transfer between devices may be necessary.
The tile-size is chosen as a function of GPU memory: each device needs to hold one block column plus a single block at any given time. An analysis of the scalability of our implementation is in Appendix~\ref{app:ooc}.

\subsection{Optimizing data transfers and other improvements.}
The speed of computations on GPUs is such that data transfers to and from the devices become significant bottlenecks. We have described earlier how, for matrix-vector products, the computed blocks of $K_{nm}$ never leave the device. 
Further, optimization is possible by parallelizing computations and
data transfers. Indeed, modern GPUs have an independent and parallel control on the following activities: loading from RAM, saving to RAM, performing computations. 
By running three parallel threads for the same GPU and assuming equal duration of each piece of work, we can run $t$ GPU computations in $t+2$ time units instead of $3t$ time units for a serial implementation (see Figure~\ref{fig:super-scalar}, where $t=3$). This guarantees near optimal usage of the GPU and in practice corresponds to a considerable speed up of matrix-vector products.\\
\noindent{\bf Leveraging the trade-off numerical precision / computational power.}
GPUs are designed to achieve peak performance with low precision floating point numbers, so much that going from 64 to 32-bit floats can correspond (depending on the exact architecture) to $\approx10\times$ throughput improvement.
However, changing precision can lead to unexpected problems. For example, computing the Gaussian kernel is commonly done by expanding the norm $\norm{\bm{x}-\bm{x}'}^2 = \bm{x}^{\top}\bm{x} - 2\bm{x}^{\top}\bm{x}' + \bm{x}'^{\top}\bm{x}'$, but in high dimensions $\norm{\bm{x}}, \norm{\bm{x}'}$ can be very large and the cross-term very negative, so their sum has fewer significant digits.
Loss of precision can lead to non positive-definite kernels causing Cholesky decomposition to fail.
To avoid this, we compute $K_{mm}$ in blocks, converting each block to 64-bit precision for the sum, and then back to 32-bits.\\
\noindent{\bf Dealing with thin submatrices.}
As a result of our block division strategies, it may happen that blocks become thin (i.e. one dimension is small). In this case, matrix operations, e.g.~using cuBLAS~\cite{cublas}, cannot leverage the full computational power. In turn this can reduce performance, breaking the inherent computational symmetry among GPUs which is crucial for the effectiveness of a parallel system like the one proposed in this paper. To guarantee good performance for this case, instead of using standard GPU operations, we perform matrix-vector products using KeOps~\cite{keops}: a specialized library to compute kernel matrices very efficiently when one dimension is small, see Table~\ref{tbl:perf-imp}.\\
\noindent{\bf Dealing with sparse datasets.}
On the other side of the spectrum, sparse datasets with high dimensionality are common in some areas of machine learning. While the kernel computed on such datasets will be dense, and thus can be handled normally, it is inefficient and in some cases impossible (e.g.~with $d\sim10^6$ as is the case for the YELP dataset we used) to convert the inputs to a dense representation.
We therefore wrapped specialized sparse linear algebra routines to perform sparse matrix multiplication~\cite{cusparse}, and adapted other operations such as the row-wise norm to sparse matrices. 
Thus our library handles sparse matrices with no special configuration, both on the GPU and -- if a GPU is not available -- on the CPU.

\section{Large-scale experiments}\label{sec:exp}\label{sec:large-exp}
We ran a series of tests to evaluate the relative importance of the computational solutions we introduced, and then performed extensive comparisons on real-world datasets.
The outcome of the first tests is given in Table~\ref{tbl:perf-imp} and is discussed in Appendix~\ref{app:rel-impact} for brevity. 
In summary, it shows a $20\times$ improvement over the base implementation of~\cite{rudi2017falkon} which runs only partially on the GPU. Such improvement is visible in equal parts for the preconditioner computations, and for the iterative CG algorithm.
For the second series of experiments we compared our implementation against three other software packages for GPU-accelerated kernel methods on several large scale datasets. All experiments were run on the same hardware, with comparable amounts of hyperparameter tuning.
Finally we compared the results of our library against a comprehensive list of competing kernel methods found in the literature.%
{
	\renewcommand{\arraystretch}{1.2}
	\begin{table}
		\caption{Relative performance improvement of the implemented optimizations w.r.t. \cite{rudi2017falkon}. The experiment was run with the HIGGS dataset, \num{1e5} centers and 10 conjugate gradient iterations.
		}%
		\label{tbl:perf-imp}
		\centering
		
			\begin{tabular}{@{}lllll@{}}
				\toprule
				Experiment & \multicolumn{2}{c}{Preconditioner} & \multicolumn{2}{c}{Iterations} \\
				\cmidrule(lr){2-3} \cmidrule(lr){4-5}
				& Time & Improvement                 & Time & Improvement \\
				\midrule
				Falkon from \cite{rudi2017falkon}         & \SI{2337}{\second} & $-$   & \SI{4565}{\second} & $-$   \\
				Float32 precision   & \SI{1306}{\second} & $1.8\times$ & \SI{1496}{\second} & $3\times$ \\
				GPU preconditioner & \SI{179}{\second}  & $7.3\times$ & \SI{1344}{\second} & $1.1\times$ \\
				2 GPUs             & \SI{118}{\second}  & $1.5\times$ & \SI{693}{\second} & $1.9\times$ \\
				KeOps              & \SI{119}{\second}  & $1\times$   & \SI{232}{\second}  & $3\times$ \\
				\midrule
				Overall improvement  &                    & $19.7\times$  &                  & $18.8\times$  \\
				\bottomrule
			\end{tabular}%
	\end{table}%
}%
We will denote our implementation by {\bf Falkon} for squared loss and by {\bf LogFalkon} for logistic loss.
Next we present the algorithms we will compare with, then shortly describe the datasets used and the experimental setting, and finally show the benchmark results. More details are in Appendix~\ref{app:experiments}.\\
{\bf Algorithms under test.}
We compare against the following software packages:
EigenPro~\cite{eigenpro2}, GPflow~\cite{GPflow2020} and GPyTorch~\cite{gpytorch18}.
The first library implements a KRR solver based on preconditioned block-coordinate gradient descent where the preconditioner is based on a truncated eigendecomposition of a data subsample. EigenPro provides a fully in-core implementation and therefore does not scale to the largest datasets we tried. On some datasets EigenPro required the training data to be subsampled to avoid GPU memory issues.
The other two packages implement several GP approximations and exact solvers, and we had to choose the model which would give a more appropriate comparison: %
we decided to avoid deep GPs~\cite{damianou13deep, wilson_deep16, cutajar17} since they share more similarities to deep nets than to kernel methods; on the other hand the exact GP -- even when implemented on GPU~\cite{gpytorch18, wang19million} -- as well as structured kernel interpolation~\cite{wilson15kissgp, gardner18product} approximations do not scale to the size of datasets we are interested in.
The only GP models which would scale up to tens of millions of points are stochastic variational GPs (SVGP).
The SVGP is trained in minibatches by maximizing the ELBO objective with respect to the variational parameters and the model hyperparameters. Stochastic training effectively constrains GPU memory usage with the minibatch size.
Hyperparameters include kernel parameters (such as the length-scale of the RBF kernel) as well as the inducing points which -- unlike in Falkon -- are modified throughout training using gradient descent. For this reason SVGP works well even with very few inducing points, and all operations can run in-core.
While GP solvers are capable of estimating the full predictive covariance, we ensured that the software did not compute it, and further we did not consider prediction times in our benchmarks. Furthermore we always considered the Gaussian kernel with a single length-scale, due to the high effort of tuning multiple length-scales for Falkon, although for GPs tuning would have been automatic.
Both GPyTorch and GPflow implement the same SVGP model, but we found the best settings on the two libraries to be different; the discrepancies in running time and accuracy between the two GP libraries come from implementation and tuning differences.
We ran all algorithms under as similar conditions as possible: 
same hardware, consistent software versions, equal floating-point precision and equal kernels (we always considered the Gaussian kernel with a single length-scale). 
Hyperparameters were optimized manually by training on a small data subset, to provide a sensible trade off between performance and accuracy: we increased the \textit{complexity} of the different algorithms until they reached high GPU utilization since this is often the knee in the time-accuracy curve. 
Details on the GP likelihoods, optimization details and other settings used to run and tune the algorithms are in Appendix~\ref{app:hyperparams}.\\
{\bf Datasets.} We used eight datasets which we believe represent a broad set of possible scenarios for kernel learning in terms of data size, data type and task ranging from MSD with $~5\times10^5$ points up to TAXI with $10^9$ points and YELP with $10^7$ sparse features.
The characteristics of the datasets are shown in table~\ref{tbl:results1} while a full description, along with details about preprocessing and relevant data splits, is available in appendix~\ref{app:datasets}.\\
{\bf Experimental setting.}
All experiments were run on a Dell PowerEdge server with 2 Intel Xeon 4116 CPUs, 2 Titan Xp GPUs and 256GB of RAM. 
Since out of the analyzed implementations only Falkon could use both GPUs effectively, we ran it both in a 2-GPU configuration (see Table~\ref{tbl:results1}) and in a single-GPU configuration (see in appendix Table~\ref{tbl:results-1gpu}) where Falkon was on average $1.6\times$ slower.
Each experiment was run 5 times, varying the random train/test data split and the inducing points.
Out of all possible experiments, we failed to run GPyTorch on TIMIT due to difficulties in setting up a multi-class benchmark (this is not a limitation of the software). Other experiments, such as EigenPro on several larger datasets, failed due to memory errors and others yet due to software limitations in handling sparse inputs (none of the examined implementations could run the sparse YELP dataset).
Finally, LogFalkon only makes sense on binary classification datasets.\\
{\bf Results.} We show the results in Table~\ref{tbl:results1}. In all cases, our library converges in less time than the other implementations: with an average speedup ranging from $6\times$ when compared to EigenPro to $>10\times$ when compared to GPyTorch.
Only on very few datasets such as AIRLINE-CLS, GPflow gets closer to Falkon's running time. 
Both models had worse accuracy than Falkon. EigenPro has generally high accuracy but can not handle large datasets at all. Finally, LogFalkon provides a small but consistent accuracy boost on binary classification problems, at the expense of higher running time.
Compared with the original Falkon library~\cite{rudi2017falkon} we report slightly higher error on HIGGS; this is attributable to the use of low-precision floating point numbers. We did not find significant performance differences for other datasets.
We defer comparisons with results from the literature to Appendix~\ref{app:more-exp}; suffice it to note that a distributed GP applied to the TAXI dataset resulted in a running-time of \SI{6000}{\second} using a system with \num{28000} CPUs~\cite{distributedGP} while we achieved similar accuracy in less time, with a much smaller computational budget.
\sisetup{detect-weight=true,detect-inline-weight=math}
{
	\begin{table}[!ht]
		\centering
		\caption{Accuracy and running-time comparisons on large scale datasets.}
		\label{tbl:results1}
		\resizebox{1\linewidth}{!}{
			\begin{threeparttable}[b]
				\begin{tabular}{l ll ll ll}
					\toprule
					& \multicolumn{2}{c}{TAXI $n\approx10^9$} &
					\multicolumn{2}{c}{HIGGS $n\approx10^7$} &
					\multicolumn{2}{c}{YELP $n\approx10^6,d\approx10^7$}  \\
					& RMSE & time & $1 - \text{AUC}$ & time & rel. RMSE & time \\
					\midrule
					
					Falkon &
					\bfseries \num{311.7 \pm 0.1}   & \bfseries \SI{3628 \pm 2}{\second} &  
					\num{0.1804 \pm 0.0003}   & \bfseries \SI{443 \pm 2}{\second}  &  
					\bfseries \num{0.810 \pm 0.001} & \bfseries \SI{1008 \pm 2}{\second} \\ 
					
					LogFalkon &
					\multicolumn{2}{c}{---} &   
					\bfseries \num{0.1787 \pm 0.0002} & \SI{2267 \pm 5}{\second} &  
					\multicolumn{2}{c}{---} \\  
					
					EigenPro &
					\multicolumn{2}{c}{FAIL} &   
					\multicolumn{2}{c}{FAIL} &   
					\multicolumn{2}{c}{FAIL} \\  
					
					GPyTorch &
					\num{315.0 \pm 0.2} & \SI{37009 (42)}{\second} &  
					\num{0.1997 \pm 0.0004} & \SI{2451 (13)}{\second} &     
					\multicolumn{2}{c}{FAIL} \\                             
					
					GPflow &
					\num{313.2 \pm 0.1} & \SI{30536 (63)}{\second} &   
					\num{0.1884 \pm 0.0003} & \SI{1174 \pm 2}{\second} &  
					\multicolumn{2}{c}{FAIL} \\  
					
					\midrule
					& \multicolumn{2}{c}{TIMIT $n\approx10^6$} &
					\multicolumn{2}{c}{AIRLINE $n\approx10^6$} &
					\multicolumn{2}{c}{MSD $n\approx10^5$} \\
					& c-error & time & rel. MSE & time & rel. error & time \\
					\midrule
					Falkon &
					\SI{32.27 \pm 0.08}{\percent} & \bfseries \SI{288 \pm 3}{\second} & 
					\bfseries \num{0.758 \pm 0.005} & \bfseries \SI{245 \pm 5}{\second}  &      
					\num{4.4834 \pm 0.0008e-3} & \bfseries \SI{62 \pm 1}{\second} \\    
					
					EigenPro &
					\bfseries \SI{31.91 \pm 0.01}{\percent} & \SI{1737 \pm 8}{\second} & 
					\num{0.785 \pm 0.005} & \SI{1471 \pm 11}{\second}\tnote{1} & 
					\bfseries \num{4.4778 \pm 0.0004e-3} & \SI{378 \pm 8}{\second} \\ 
					
					GPyTorch &
					\multicolumn{2}{c}{---} & 
					\num{0.793 \pm 0.005} & \SI{2069 \pm 50}{\second} &  
					\num{4.5004 \pm 0.0010e-3} & \SI{502 \pm 2}{\second}\\ 
					
					GPflow &
					\SI{33.78 \pm 0.14}{\percent} & \SI{2672 \pm 10}{\second} &  
					\num{0.782 \pm 0.005} & \SI{1297 \pm 2}{\second} &  
					\num{4.4986 \pm 0.0005e-3} & \SI{525 \pm 5}{\second} \\ 
					
					\midrule
					& \multicolumn{2}{c}{AIRLINE-CLS $n\approx10^6$} & \multicolumn{2}{c}{SUSY $n\approx10^6$} & & \\
					& c-error & time & c-error & time & &  \\
					\midrule
					
					Falkon &
					\SI{31.5 \pm 0.2}{\percent} & \bfseries \SI{186 \pm 1}{\second} &   
					\SI{19.67 \pm 0.02}{\percent} & \bfseries \SI{22(0)}{\second} & & \\ 
					
					LogFalkon &
					\bfseries \SI{31.3 \pm 0.2}{\percent} & \SI{1291 \pm 3}{\second} & 
					\bfseries \SI{19.58 \pm 0.03}{\percent} & \SI{83 \pm 1}{\second} & & \\ 
					
					EigenPro &
					\SI{32.5 \pm 0.2}{\percent} & $\SI{1629 \pm 1}{\second}^1$ & 
					\SI{20.08 \pm 0.55}{\percent} & $\SI{90(0)}{\second}^2$ & & \\ 
					
					GPyTorch &
					\SI{32.5 \pm 0.2}{\percent} & \SI{1436 \pm 2}{\second} &  
					\SI{19.69 \pm 0.03}{\percent} & \SI{882 \pm 9}{\second} & & \\ 
					
					GPflow &
					\SI{32.3 \pm 0.2}{\percent} & \SI{1039 \pm 1}{\second} & 
					\SI{19.65 \pm 0.03}{\percent} & \SI{560 \pm 11}{\second} & & \\ 
					
					\bottomrule
				\end{tabular}
			\end{threeparttable}
		}\\
		{\footnotesize $^1$Using a random subset of \num{1e6} points for training. $^2$Using a random subset of \num{6e5} points for training.}
		\vspace{-0.3cm}
	\end{table}
}

\section{Conclusions}
Making flexible and easy to use machine learning libraries available is one of the keys of the recent success of machine learning.
Here, we contribute to this effort by developing a library for large scale kernel methods. We translate algorithmic ideas into practical solutions,  
using a number of carefully design computational approaches specifically adapted to the GPU. 
The resulting library achieves excellent performance both in terms of accuracy and computational costs. 
A number of further developments are possible building on our work. For example, considering other loss functions or optimization approaches, and especially more structured kernels~\cite{hierachical17} that could further improve efficiency. 

\section*{Acknowledgments}
This material is based upon work supported by the Center for Brains, Minds and Machines (CBMM), funded by NSF STC award CCF-1231216, and the Italian Institute of Technology. We gratefully acknowledge the support of NVIDIA Corporation for the donation of the Titan Xp GPUs and the Tesla k40 GPU used for this research.
Part of this work has been carried out at the Machine Learning Genoa (MaLGa) center, Università di Genova (IT)
L. R. acknowledges the financial support of the European Research Council (grant SLING 819789), the AFOSR projects FA9550-17-1-0390  and BAA-AFRL-AFOSR-2016-0007 (European Office of Aerospace Research and Development), and the EU H2020-MSCA-RISE project NoMADS - DLV-777826. 
This work was funded in part by the French government under management of Agence Nationale de la Recherche as part of the ``Investissements d’avenir'' program, reference ANR-19-P3IA-0001 (PRAIRIE 3IA Institute).

\bibliographystyle{plainnat}
\bibliography{main}

\clearpage
\appendix
\section{Further experiment details and results}\label{app:experiments}

\subsection{Relative impact of performance optimizations}\label{app:rel-impact}

We performed an experiment to analyze how much improvement was due to the different performance optimization steps. We ran Falkon on the HIGGS dataset several times with the same hyperparameters ($m=\num{1e5}$ and \num{10} epochs), but with different \textit{features} enabled. Each feature roughly corresponds to one of the performance optimizations discussed in Section~\ref{sec:methods}.
Our baseline model is very similar to the original Falkon implementation~\cite{rudi2017falkon}, where the preconditioner ran on the CPU, float64 precision was being used, but matrix-vector multiplications for the CG algorithm were GPU accelerated.
As a first optimization we used float32 precision for all computations, with care taken to avoid errors in the Cholesky decomposition as discussed in Section~\ref{sec:methods}. This immediately resulted in a $2\times$ speedup for the CPU part, and $3\times$ for the GPU part.
Switching to a GPU preconditioner (using the algorithms described in Appendix~\ref{app:ooc}) gave a huge boost to the preconditioner running time which went from more than \SI{20}{\minute} to just under \SI{3}{\minute}. 
Adding a second GPU produced a perfect $2\times$ speedup for the CG iterations, and a more modest $1.5\times$ speedup for the preconditioner which \begin{enumerate*}[label=\alph*)]\item involves operations which are not perfectly parallelizable and \item incurs in some fixed startup costs\end{enumerate*}. 
Finally, since the HIGGS dataset has only 9 features (thus the data matrix is thin), we can use KeOps~\cite{keops} with great benefits to the speed of matrix-vector multiplications.
Overall our implementation provides a nearly $20\times$ improvement over the baseline, which makes learning on several huge datasets doable in a matter of minutes.

\subsection{Multi-GPU scalability}\label{app:multi-gpu-exp}

\begin{figure}[!ht]
	\centering
	\includegraphics[width=0.5\linewidth]{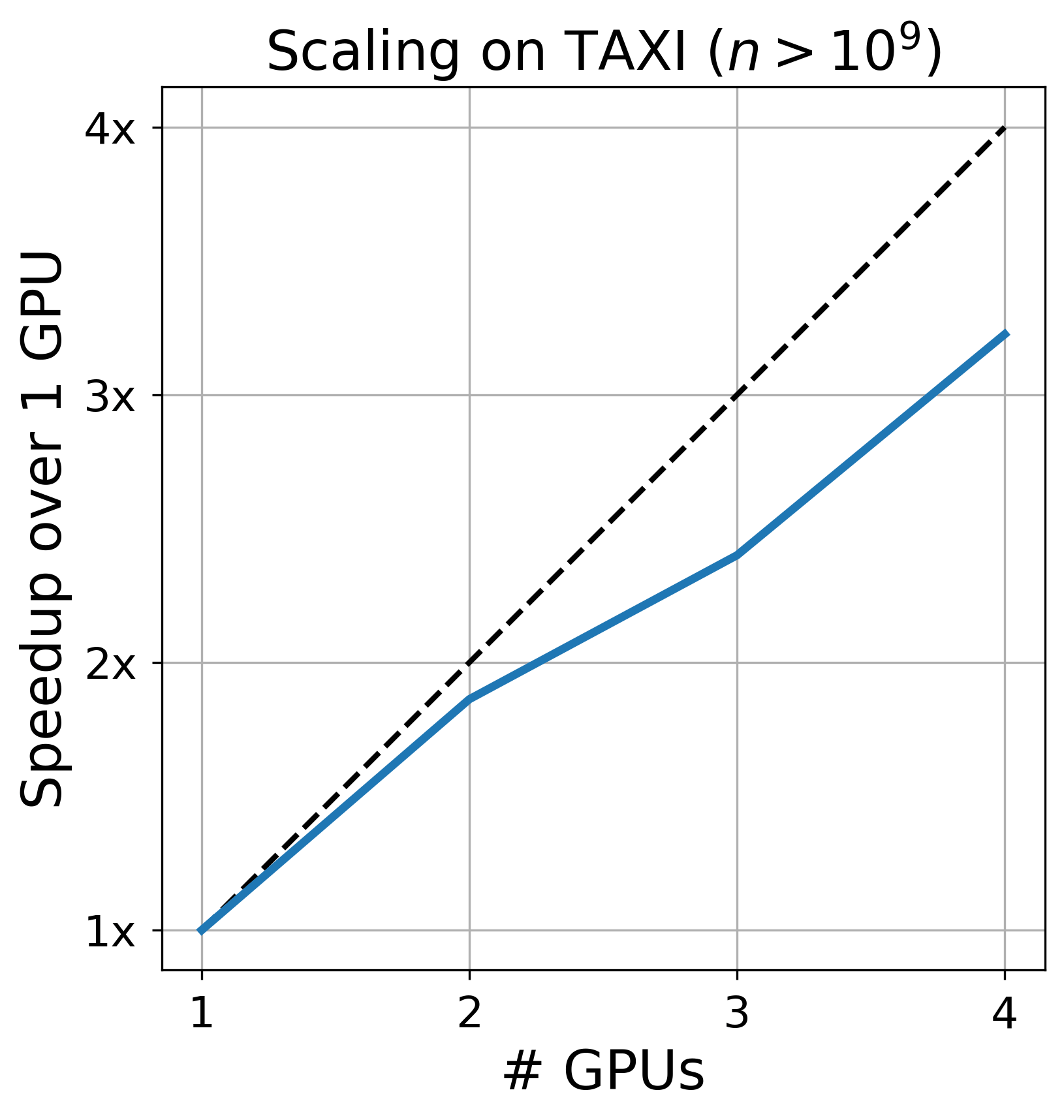}
	\caption{Multi-GPU scalability of Falkon on the TAXI dataset (settings are the same as per Table~\ref{tbl:hparams}). Falkon scales remarkably well, with even 4 GPUs.}
	\label{fig:4gpu}
\end{figure}

In this section we look into the scalability of our implementation across multiple GPUs. Scalability results for the full Falkon algorithm on the TAXI dataset are shown in Figure~\ref{fig:4gpu}. This result depends on scaling both the preconditioner and the conjugate gradient iterations.
The preconditioner itself is computed with three main operations: two Cholesky decompositions and one triangular matrix multiplication (this is called the LAUUM operation in LAPACK terms), see Figure~\ref{fig:preconditioner} for more details. 
Each CG iteration instead consists of two multiplications between the kernel matrix and an arbitrary vector. 
First we look at the scalability of the preconditioner operations with multiple GPUs. Then we examine our out-of-core matrix-vector product implementation and compare it to KeOps for different settings of $n$ and $d$.

\paragraph{Preconditioner scalability.}

\begin{figure}
	\centering
	\subcaptionbox{Parallel LAUUM.\label{fig:par-lauum}}{%
		\includegraphics[width=0.49\linewidth]{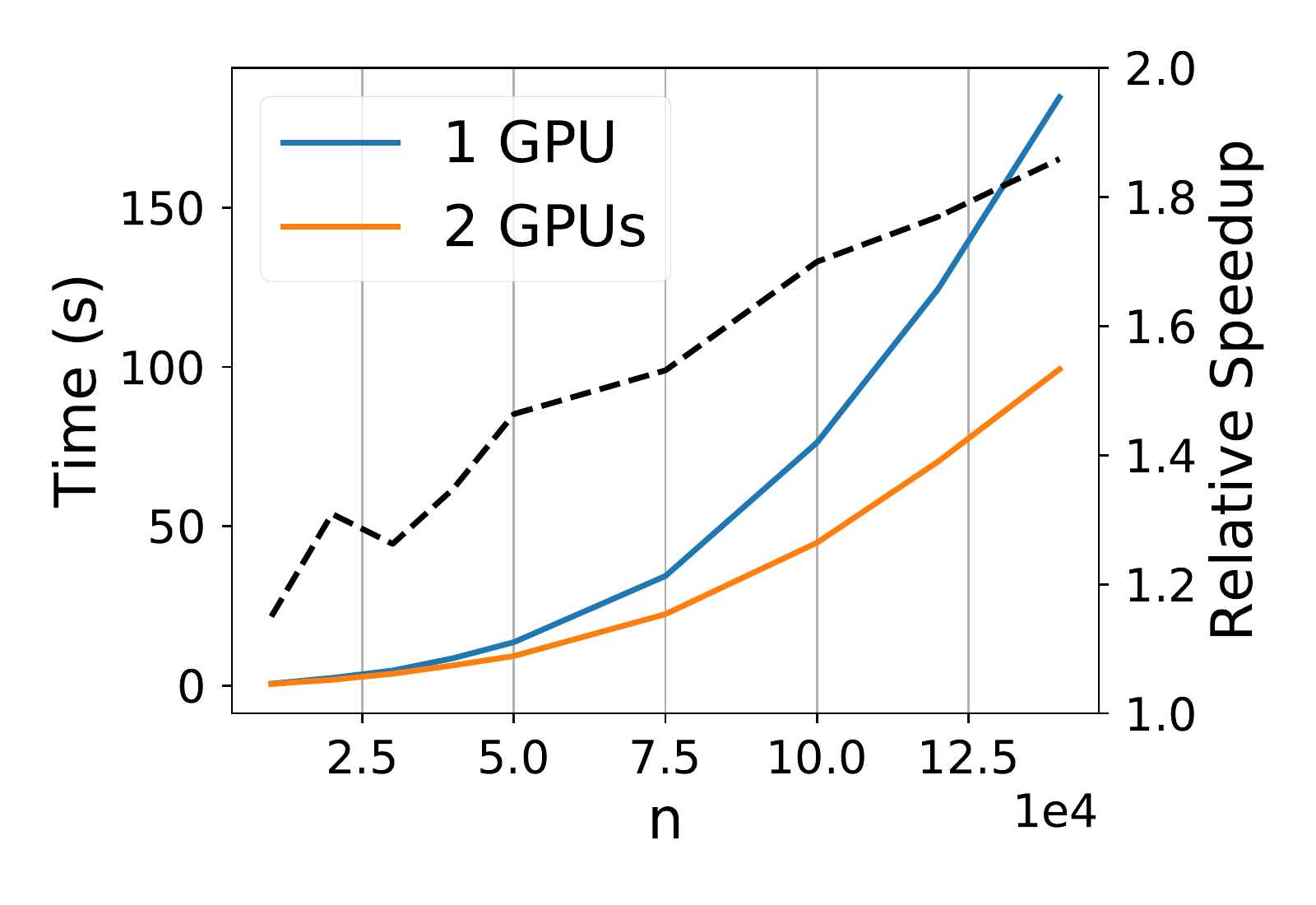}%
	}\hfill
	\subcaptionbox{Parallel Cholesky decomposition.\label{fig:par-potrf}}{%
		\includegraphics[width=0.49\linewidth]{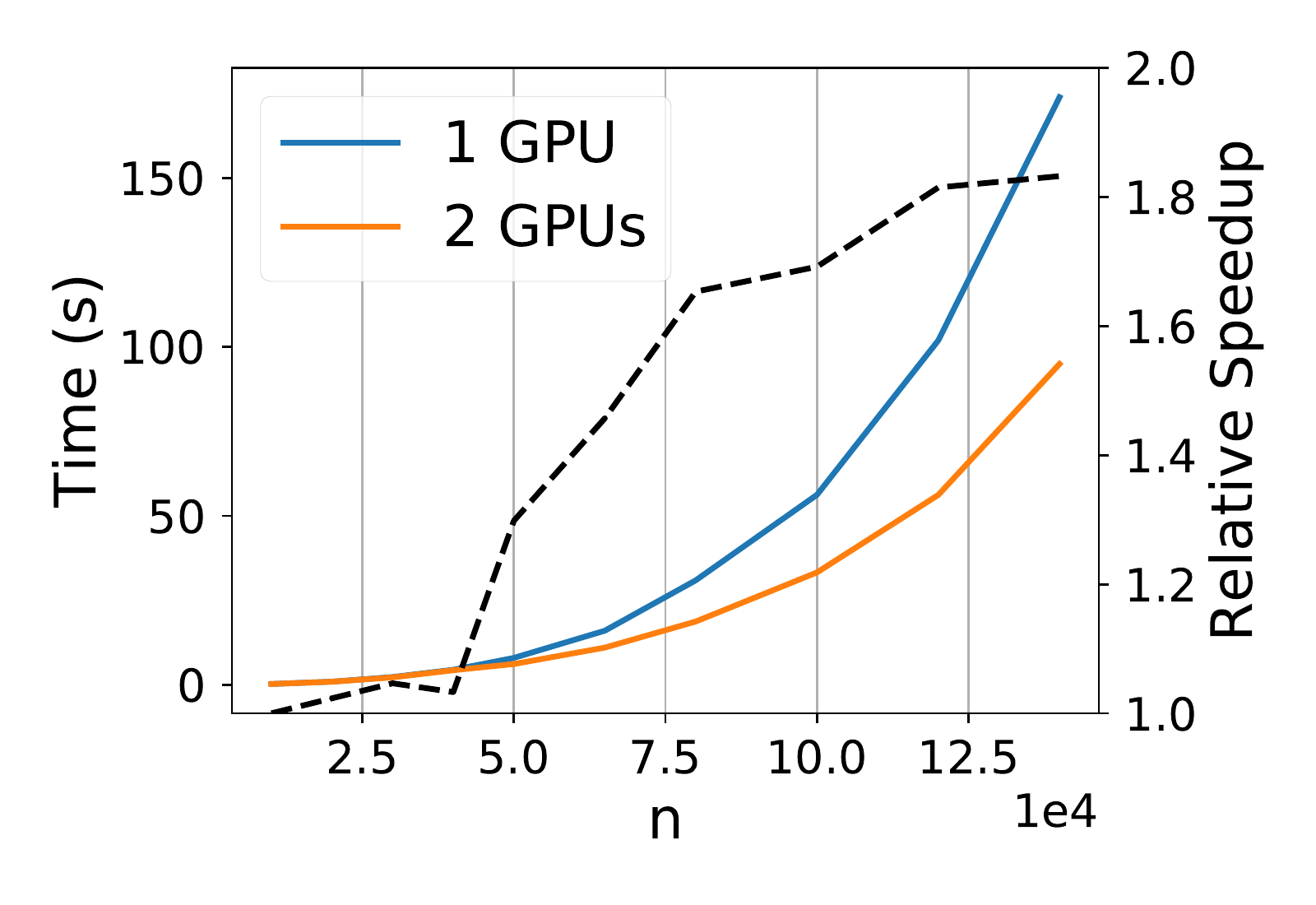}%
	}%
	\caption{Running time of two preconditioner operations with one and two GPUs. The relative speed-up with 2 GPUs is shown in the black dashed line. The LAUUM operation (triangular matrix multiplication) was run out-of-place, which is theoretically easier to parallelize, while the Cholesky decomposition was run in-place.}\label{fig:par-precond}
\end{figure}

Figure~\ref{fig:par-precond} shows the results from running both triangular matrix multiplication and the Cholesky decomposition with one and two GPUs. At low matrix sizes the speedup with two GPUs is negligible, especially for the Cholesky decomposition. In such cases it is best to use a single GPU (especially since for $n=40000$ the whole matrix fits in GPU memory, so an in-place decomposition can be used). 
With higher matrix sizes, having more than one GPU starts bringing real benefits, with a peak speedup around $1.8\times$ for preconditioners of size \num{140000}.
The factors blocking such speedup from reaching a perfect $2\times$ are different for the two operations. 
Since the LAUUM operation was run out-of-place (see Appendix~\ref{app:ooc} for more details), it does not need any synchronization -- and should therefore be able to scale well across multiple GPUs. The main blocking factor is the operation at Line~\ref{line:lauum} of Algorithm~\ref{alg:oocLauum} which is executed on the CPU (since an equivalent implementation does not exist in cuSOLVER), thus both GPU threads must share the same CPU resources. We left porting the LAUUM operation to the GPU as future work, but it has the potential to speed up the LAUUM operation considerably.
For the Cholesky decomposition the limiting factors are the data-dependencies intrinsic to the algorithm which cannot be easily solved.

\begin{figure}
	\centering
	\begin{subfigure}{.5\textwidth}
		\centering
		\includegraphics[width=0.9\textwidth]{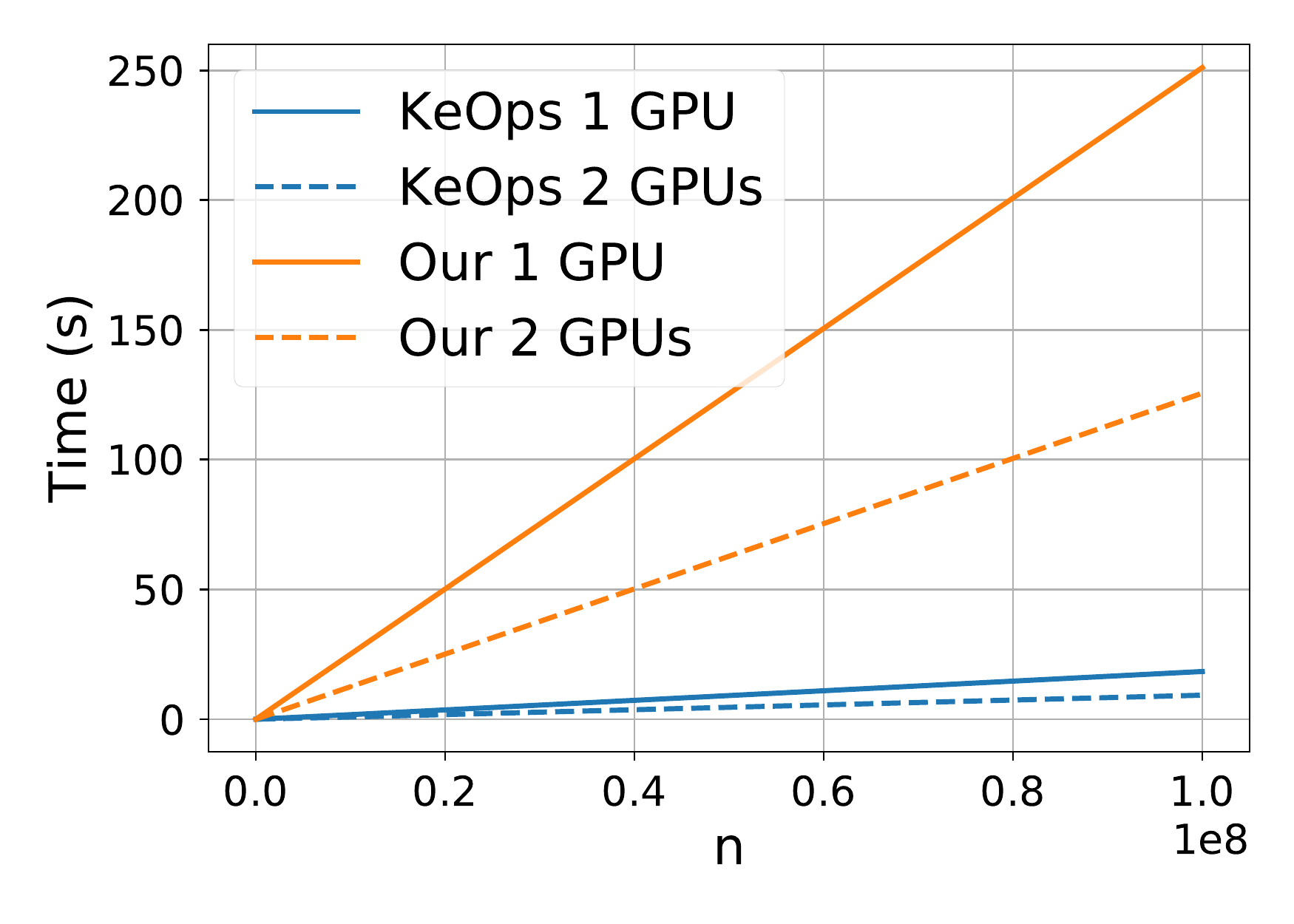}
		\caption{}
		\label{subfig:mvm_impl_n}%
	\end{subfigure}%
	\begin{subfigure}{.5\textwidth}
		\centering
		\includegraphics[width=0.9\textwidth]{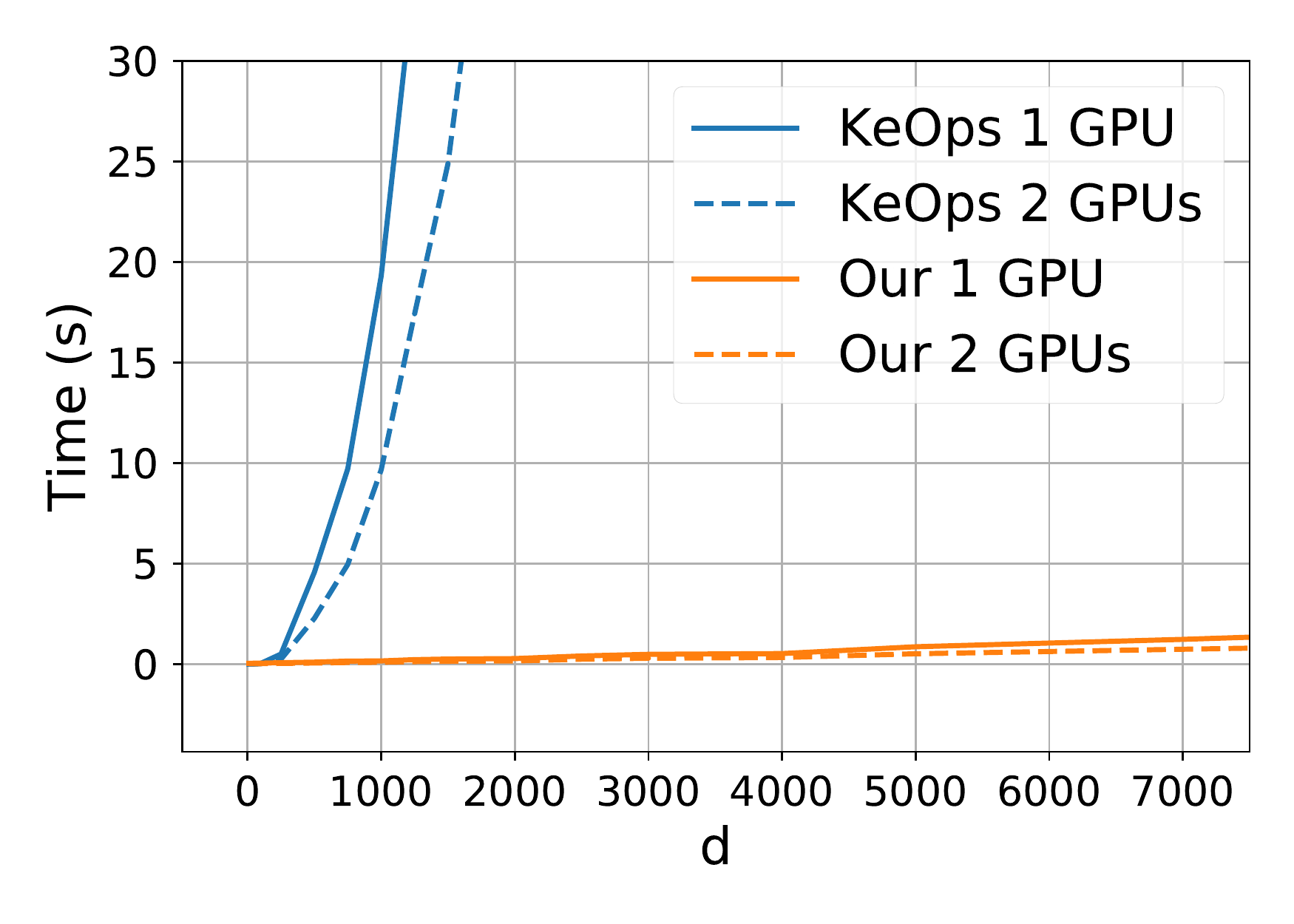}
		\caption{}
		\label{subfig:mvm_impl_d}%
	\end{subfigure}%
	\caption{
		Scaling of matrix-vector implementations where the matrix is the Gaussian kernel. In~\protect\subref{subfig:mvm_impl_n} we have set $m=\num{20000}$, $d=10$ and $n$ is variable; 
		in~\protect\subref{subfig:mvm_impl_d} we set $m=n=\num{20000}$ and we vary $d$. All experiments are run on 1 and 2 GPUs on single precision random data.
	}
	\label{fig:mvm_impl}
\end{figure}

\paragraph{Comparing different MVM implementations.}
We compare our specialized routine for the kernel-vector multiplication $k(X^{(1)}, X^{(2)}) \bm{v}$ implemented in Python, leveraging PyTorch for GPU computations, against the native CUDA implementation from KeOps~\cite{keops}. Using a similar notation for the dimensions as in the main text we have $X^{(1)} \in \real{n\times d}, X^{(2)} \in \real{m \times d}, \bm{v} \in \real{m \times 1}$ and $k(\cdot, \cdot)$ is a kernel function.
Two distinct scenarios arise in different settings: increasing the number of data points $n$ produces linear scaling for both implementations, with KeOps being approximately 10 times faster than our implementation (see Figure~\ref{subfig:mvm_impl_n}).
Increasing the data dimensionality $d$ our implementation scales linearly, but KeOps scales polynomially, so as it is obvious from Figure~\ref{subfig:mvm_impl_d} KeOps can not be used when the data is high-dimensional. A caveat of this plot is that KeOps is continuously evolving, and is likely to improve performance with large $d$ in the future.
In our final algorithm we set a threshold on the data dimensionality and switch implementation based on this.
Finally note that this operation scales almost perfectly with multiple GPUs.

\subsection{Additional information on the datasets}\label{app:datasets}
We used several datasets which we believe represent a broad set of scenarios for kernel learning, in terms of data size, data type, and learning task. We normally used a standard random split with 80\% training, 20\% testing data unless predefined splits existed (as noted below). Preprocessing mostly consisted in basic data cleaning and data standardization to zero mean and unit standard deviation; we comment in more detail below on specific preprocessing steps applied to the individual datasets.

\paragraph{HIGGS} has dimensions $n = 1.1 \times 10^7, d = 28$ and a binary target. It was preprocessed to 0 mean and unit variance. Results are reported on a 80-20 split with $1$ minus the AUC metric in Table~\ref{tbl:results1} and with the binary classification error in Table~\ref{tbl:results2}. It is available for download at \url{https://archive.ics.uci.edu/ml/datasets/HIGGS}.
\paragraph{TIMIT} has dimensions $n = 1.2 \times 10^6, d = 440$ and a multiclass target with $144$ classes. TIMIT comes from audio data, and our dataset uses the \SI{10}{\milli\second} resampling rate as in~\cite{eigenpro17,eigenpro2}. It was preprocessed to 0 mean and unit standard deviation. The error metric is classification error on a subset of classes (as used in~\cite{eigenpro17}), and is calculated over a standardized subset of $\num{57242}$ samples. It is available for download at \url{https://catalog.ldc.upenn.edu/LDC93S1}.
\paragraph{YELP} has dimensions $n = 1.5 \times 10^6, d = 6.52 \times 10^7$ and a continuous target.
This dataset consists of text reviews, labeled with their star rating. We used the same data as~\cite{Yelp16} (Yelp round 9 dataset), processed by extracting all 3-grams and encoding each review by a count vector which tells us which 3-grams are present. Such encoding produces a large number of sparse features which is reflected in the huge dimensionality of this dataset. Since the data is sparse we did not normalize it. The error metric is RMSE, calculated on random 20\% of the samples. The dataset can be provided on request.
\paragraph{TAXI} has dimensions $n = 1.1 \times 10^9, d = 9$ with a continuous target. Data are normalized to have zero-mean and unit standard deviation; reported error is RMSE on a 20\% random sub-sample. The data can be downloaded by following instructions at \url{https://github.com/toddwschneider/nyc-taxi-data}. Consistently with other users of this dataset~\cite{distributedGP} we took the data from January 2009 to December 2015, excluding outliers (taxi trips more than 5 hours long) and trips where the pickup or drop off location is outside of NYC.
\paragraph{AIRLINE} has dimensions $n = \num{5.93e6}, d=8$ and a continuous target. Data are normalized to zero-mean and unit standard deviation, and the error is the MSE over normalized targets calculated on random test-sets of size \SI{33}{\percent} of the full data (consistently with the literature~\cite{vffgp_hensman17,hensman13}).
The same dataset is also used for binary classification by thresholding the target at 0, which results in the \textbf{AIRLINE-CLS} dataset.
For this latter variation we used \num{100000} random points for testing, reporting classification error in Table~\ref{tbl:results1} and $1$ minus the AUC in Table~\ref{tbl:results2} to facilitate comparisons with the literature.
The data can be downloaded from \url{https://www.transtats.bts.gov/Fields.asp?Table_ID=236} and \url{http://stat-computing.org/dataexpo/2009/supplemental-data.html}.
\paragraph{MSD} has dimensions $n = \num{5.1e5}, d = 90$ with continuous target. Data are normalized to zero-mean and unit standard deviation, and we report the relative error over a standard test-set of size \num{51630}. The dataset can be downloaded from \url{https://archive.ics.uci.edu/ml/datasets/YearPredictionMSD}.
\paragraph{SUSY} has dimensions $n= \num{5e6}, d=18$ with binary target. Data are normalized to zero-mean and unit standard deviation. We report the classification error on 20\% of the data. Data is available from the UCI repositories \url{https://archive.ics.uci.edu/ml/datasets/SUSY}.

\subsection{Additional information on the experimental settings}\label{app:hyperparams}

\begin{enumerate}
	\item EigenPro2.
	Its only hyperparameters -- other than the kernel parameters -- are the ones governing the preconditioner's complexity making EigenPro easy to tune.
	It is however limited to datasets which fit entirely in GPU memory, so can not easily scale to larger datasets; to alleviate this problem, consistently with the original paper, some experiments were run on sub-sampled datasets.
	Furthermore, on some experiments we found it necessary to manually tune the learning rate (we divided the automatically inferred learning rate by a fixed integer, denoted by $\eta\div$ in Table~\ref{tbl:hparams}).
	
	\item GPFlow (v2.1.3).
	We used the SVGP model with Gaussian likelihood for regression, Bernoulli for binary classification and Softmax for multi-class problems.
	We used Adam for optimization and tuned the learning rate, the number of inducing points, and the constraints on the variational distribution covariance (i.e. diagonal or full covariance matrix).
	We found that using a full covariance matrix was rarely beneficial and increased training times slightly, so all final experiments used a diagonal covariance matrix. The number of parameters was $m\times d + m\times2 + 3$ which includes the inducing points, the variational parameters, two parameters for the Gaussian kernel (lengthscale and variance) and the variance of the likelihood. For multi-class problems separate variational parameters were trained for each class.
	Since we wished to use single-precision floating point numbers in order to make GPU training more efficient, we found that natural gradient optimization was unstable. It remains to be seen whether the tradeoff between double-precision data and natural gradient optimization could further improve results. We further tested the benefits of using whitening of the inducing points, and found that it decreased per-epoch running times by about $2\times$, while at the same time slowing down convergence by around the same amount. In practice this meant that the difference in global running time was not strongly affected by whitening, and we ended up using it only for the HIGGS data.
	
	\item GPyTorch (v.1.2.0). We used the SVGP model with Gaussian and Bernoulli likelihoods.
	We were unable to run GPyTorch's SVGP model on the TIMIT dataset due to problems in dealing with multiple outputs. We used the natural gradient optimizer to learn the variational parameters, and Adam to learn the other hyperparameters. The learning rate of the two optimizers was kept equal and tuned for best performance. We further optimized the number of inducing points, and variational distribution constraints. In practice we found that we had to use the natural gradient variational distribution for regression problem, and the lower-triangular parametrization for classification problems (which are non-conjugate). We additionally tested whether whitening the inducing points was beneficial: in practice we found that using the unwhitened strategy was around $3\times$ faster and did not hamper convergence, so we selected it for all experiments.
	While GPyTorch is theoretically able to run on multi-GPU systems, we noticed that this feature was not available for the SVGP model thus we always used a single GPU; furthermore, while a KeOps integration into GPyTorch is available, we found that for the SVGP model it would increase the running time, so we did not use it.
	The trained parameters were the same as for GPFlow plus another scalar for the GP mean.
	
	\item Falkon. We tuned the kernel length-scale, number of inducing points and regularization amount. We used a coarse to fine approach to tune the length-scale which gives good results with a limited number of validation runs.
	
	\item Logistic Falkon. Here we tuned the kernel length-scale, number of inducing points and regularization path. We found that the algorithm is not very sensitive to the exact regularization path: it is sufficient to set the final $\lambda$, and many different paths which lead to such value will work in the same way.
\end{enumerate}

\begin{table}
	\caption{Summary of the most important hyperparameter settings for all algorithm-dataset combinations. We denote by $\eta$ the learning rate, by \textit{subsample} the amount of training-set subsampling that was performed (i.e. training was done on a smaller dataset), and by Newton steps the number of separate runs of the main Falkon algorithm for Logistic Falkon (see Appendix~\ref{app:logistic}).}
	\label{tbl:hparams}
	\centering
	\resizebox{0.97\linewidth}{!}{
		\begin{tabular}{llllllllll}
			\toprule
			& & AIRLINE & AIRLINE-CLS & MSD & SUSY & TIMIT & YELP & HIGGS & TAXI \\
			\midrule
			& n & \num{5.93e6} & \num{5.93e6} & \num{5.1e5} &\num{5e6} &\num{1.2e6} &\num{1.6e6} &\num{11e7}   &\num{1.15e9} \\
			& d &\num{8}       & \num{8}      & \num{90}    &\num{18}  &\num{440}   &\num{6.5e7} &\num{28}     &\num{9} \\
			& labels & reg     & 2-cls        & reg         & 2-cls    & 144-cls    & reg        & 2-cls       & reg \\
			\midrule
			
			\multirow[t]{4}{*}{Falkon}
			& m         & \num{1e5}  & \num{1e5}  & \num{5e4}  & \num{3e4}  & \num{1e5}  & \num{5e4}  & \num{1.2e5} & \num{1e5}  \\
			& $\sigma$  & \num{0.9}  & \num{0.9}  & \num{7}    & 3          & \num{14.5} & \num{20}   & \num{3.8}   & \num{1}  \\
			& $\lambda$ & \num{1e-8} & \num{1e-8} & \num{2e-6} & \num{1e-6} & \num{5e-9} & \num{1e-6} & \num{3e-8}  & \num{2e-7} \\
			& epochs    & 20         & 10         & 10         & 5          & 5          & 10         & 10          & 7 \\
			
			\multirow[t]{4}{*}{LogFalkon}
			& m           & --  & \num{1e5}  & -- & \num{2e4}  & -- & -- & \num{1e5}  & --  \\
			& $\sigma$    & --  & \num{0.9}  & -- & 3          & -- & -- & \num{5}    & --  \\
			& $\lambda$   & --  & \num{1e-9} & -- & \num{1e-8} & -- & -- & \num{1e-9} & --  \\
			& Newt. steps & --  & 9          & -- & 6          & -- & -- & 9          & --  \\
			
			\multirow[t]{3}{*}{GPyTorch}
			& m      & 2000       & 2000       & 3000       & 2000       & -- & -- & 2000       & 1000 \\
			& $\eta$ & \num{5e-3} & \num{2e-3} & \num{2e-3} & \num{1e-3} & -- & -- & \num{2e-2} & \num{2e-3} \\
			& epochs & 20         & 20         & 20         & 20         & -- & -- & 15         & 5 \\
			
			\multirow[t]{3}{*}{GPflow}
			& m      & 2000       & 2000       & 3000       & 2000       & 2000       & -- & 2000       & 1000       \\
			& $\eta$ & \num{5e-3} & \num{5e-3} & \num{2e-3} & \num{3e-3} & \num{1e-2} & -- & \num{2e-2} & \num{3e-3} \\
			& epochs & 25         & 20         & 45         & 10         & 15         & -- & 60         & 10         \\
			& whiten & no         & no         & no         & no         & no         & -- & yes        & no         \\
			
			\multirow[t]{3}{*}{EigenPro}
			& $\eta\div$ & 10        & 12        & 20 & 1         & 1  & -- & -- & -- \\
			& subsample  & \num{1e6} & \num{1e6} & -- & \num{6e5} & -- & -- & -- & -- \\
			& epochs     & 9         & 10        & 9  & 1         & 4  & -- & -- & -- \\
			\bottomrule
		\end{tabular}
	}
\end{table}

\subsection{Additional benchmarks}

In Table~\ref{tbl:results-1gpu} we show the performance of the Falkon algorithm on all considered datasets for 1 and 2 GPUs side by side. It is clear that larger datasets scale better with more GPUs since the startup cost (mostly taken up by CUDA initialization) and the lower scaling ratio of the preconditioner are amortized.
{
	\renewcommand{\arraystretch}{1.2}
	\begin{table}
		\centering
		\caption{Benchmark timings using a single GPU. The relative slowdown with respect to Falkon on 2 GPUs is also provided for comparison with Table~\ref{tbl:results1}. }
		\label{tbl:results-1gpu}
		\begin{tabular}{llll}
			\toprule
			& 1 GPU & 2 GPUs & Relative change \\
			\midrule
			TAXI        & \SI{7215 \pm 4}{\second} & \SI{3628 \pm 2}{\second} & $1.99\times$ \\
			HIGGS       & \SI{715 \pm 6}{\second}  & \SI{443 \pm 2}{\second}  & $1.61\times$ \\
			YELP        & \SI{1981 \pm 6}{\second} & \SI{1008 \pm 2}{\second} & $1.97\times$ \\
			TIMIT       & \SI{416 \pm 4}{\second}  & \SI{288 \pm 3}{\second}  & $1.44\times$ \\
			AIRLINE     & \SI{334 \pm 2}{\second}  & \SI{245 \pm 5}{\second}  & $1.36\times$ \\
			MSD         & \SI{81(0)}{\second}      & \SI{62 \pm 1}{\second}   & $1.31\times$ \\
			AIRLINE-CLS & \SI{391 \pm 5}{\second}  & \SI{269 \pm 3}{\second}  & $1.45\times$ \\
			SUSY        & \SI{29 \pm 1}{\second}   & \SI{22(0)}{\second}      & $1.32\times$ \\
			\bottomrule%
		\end{tabular}
	\end{table}
}

In Table~\ref{tbl:results-svm} we compare the running times of Falkon and ThunderSVM~\cite{wenthundersvm18} on three popular image datasets. ThunderSVM was chosen among several SVM implementations as it runs entirely on the GPU, and can thus solve the hinge-loss problem quickly for problems of moderate size. Smaller datasets than the ones used for previous experiments were considered, since ThunderSVM solves the full SVM problem and thus suffers from cubic time scaling. The results obtained show that Falkon can work efficiently even on smaller datasets, resulting between 2 and 10 times faster than ThunderSVM (depending on problem size), with comparable accuracy. 
To further shave off some time, we implemented a version of Falkon which runs entirely inside the GPU: we call this \textbf{InCoreFalkon}, and it can only be used on smaller datasets which fit inside the GPU, leaving some space to spare which is used for the preconditioner and the other computations. Table~\ref{tbl:results-svm} shows that InCoreFalkon gives a further speed-up of -- on average -- $2\times$ compared to the standard implementation.

\begin{table}
	\renewcommand{\arraystretch}{1.2}
	\centering
	\caption{Comparing the running times of Falkon, the in-core version of Falkon and ThunderSVM on three image datasets. Hyperparameters (especially the number of inducing points $m$) were tuned so that the two algorithms obtained approximately the same accuracy.}
	\label{tbl:results-svm}
	\begin{tabular}{lp{9em}p{9em}p{9em}}
		\toprule
		& MNIST \newline $n=6\cdot 10^4, d=780$ &  CIFAR10\newline $n=6\cdot 10^4, d=1024$ & SVHN\newline $n=7\cdot 10^4, d=1024$ \\
		\midrule
		Falkon       & \SI{10.9}{\second} & \SI{13.7}{\second} & \SI{17.2}{\second}   \\
		InCoreFalkon & \SI{6.5}{\second} & \SI{7.9}{\second} & \SI{6.7}{\second}    \\
		ThunderSVM   & \SI{19.6}{\second} & \SI{82.9}{\second} & \SI{166.4}{\second}  \\
		\bottomrule
	\end{tabular}
\end{table}

\subsection{Performance comparisons in a literature review}\label{app:more-exp}
{
	\renewcommand{\arraystretch}{1.5} 
	\begin{table}
		\caption{Survey of results on the datasets we considered, as reported in the literature. We report the result of our implementation (Falkon) next to other implementations, along with the time taken and the hardware used (where available).}
		\label{tbl:results2}
		\centering
		\resizebox{1\linewidth}{!}{
			\begin{tabular}{lll|lp{9em}p{11em}}
				\toprule
				Dataset & \multicolumn{2}{c}{Falkon} & \multicolumn{3}{c}{Other methods} \\
				\cmidrule(r){2-3} \cmidrule(r){4-6}
				& error & time & error & time & reference \\
				\midrule
				
				\multirowcell{2}{TAXI \\ (metric: RMSE)} &
				\multirow[t]{1}{*}{$\num{311.7 \pm 0.1}$} & \multirow[t]{1}{*}{$\SI{3628 \pm 2}{\second}$} &
				\num{309.7} & \SI{6000}{\second} \newline \num{28000} vCPUs (AWS) & ADVGP~\cite{distributedGP} \\
				
				\cmidrule(r){2-6}
				\multirowcell{2}{HIGGS \\ (metric: c-err)} &
				\multirow[t]{1}{*}{\SI{25.78 \pm 0.03}{\percent}} & \multirow[t]{1}{*}{$\SI{443 \pm 2}{\second}$} &
				\SI{32.87}{\percent} & $\SI{1392}{\second}$ \newline on 14 node cluster & liquidSVM~\cite{liquisvm17} \\
				
				\cmidrule(r){2-6}
				\multirowcell{4}{YELP \\ (metric: RMSE)} &
				\multirow[t]{1}{*}{\num{0.810 \pm 0.001}} & \multirow[t]{1}{*}{\SI{1008 \pm 2}{\second}} & 
				\num{0.861} & $\approx \SI{3500}{\second}$ & Nystr{\"o}m~\cite{Yelp16} \\
				& & & \num{0.854} & $\approx \SI{30000}{\second}$ \newline on 128 machines (AWS) & Full linear kernel~\cite{Yelp16} \\
				
				\cmidrule(r){2-6}
				\multirowcell{4}{AIRLINE \\ (metric: MSE)} &
				\multirow[t]{1}{*}{$\num{0.758 \pm 0.005}$} & \multirow[t]{1}{*}{$\SI{245 \pm 5}{\second}$} &
				$\num{0.827\pm0.004}$ & $\SI{265\pm 6}{\second}$ \newline on a laptop & VFF-GP~\cite{vffgp_hensman17} \\
				& & & $\num{0.791 \pm 0.005}$ & $\SI{18360 \pm 360}{\second}$ \newline on a cluster & SVIGP~\cite{vffgp_hensman17} \\
				
				\cmidrule(r){2-6}
				\multirowcell{4}{MSD \\ (metric: rel. err.)} &
				\multirow[t]{2}{*}{\num{4.48e-3}} & \multirow[t]{2}{*}{$\SI{62 \pm 1}{\second}$} &
				$\approx \num{4.55e-3}$ & \SI{210}{\second} \newline on IBM POWER8 & Hierarchical~\cite{hierachical17} \\
				& & & \num{4.58e-3} & \SI{289}{\second} \newline on 8 r3.8xlarge (AWS) & Faster KRR~\cite{fasterkrr17} \\
				
				\cmidrule(r){2-6}
				\multirowcell{5}{\shortstack[l]{AIRLINE-CLS \\ (metric: AUC)}} &
				\multirow[t]{5}{*}{\num{0.739 \pm 0.002}} & \multirow[t]{5}{*}{\bfseries \SI{186 \pm 1}{\second}} &
				$\num{0.781\pm0.001}$ & $\SI{14328}{\second}$ & Varitional Deep GP~\cite{wilson_deep16}\\
				& & & $\num{0.694}$ & $\SI{5200}{\second}$ & TT-GP~\cite{tensor_train18} \\
				& & & $\num{0.788}$ & $\SI{1375}{\second}$ & Deep TT-GP~\cite{tensor_train18} \\
				& & & \num{0.665} & $\SI{80000}{\second}$ & cVGP\cite{classif_vgps15} \\
				& & & \num{0.785} & $\approx \SI{5000}{\second}$ & RF Deep GPs~\cite{cutajar17} \\
				
				\cmidrule(r){2-6}
				\multirowcell{2}{SUSY \\ (metric: c-err)} &
				\multirow[t]{2}{*}{\SI{19.67 \pm 0.02}{\percent}} & \multirow[t]{2}{*}{\bfseries \SI{22(0)}{\second}} &
				$\approx 20\%$ & $\approx \SI{2000}{\second}$ \newline on IBM POWER8 & Hierarchical~\cite{hierachical17} \\
				& & & $19.8\%$ & \SI{58}{\second} \newline on 1 Titan Xp & EigenPro 2.0~\cite{eigenpro2} \\
				
			\end{tabular}
		}
	\end{table}
}

We scanned the literature for results which used kernel methods on the datasets considered in this paper, which reported both accuracy and running times. This allowed us to confirm that the results reported in our benchmarks (see Table~\ref{tbl:results1}) were in-line with what had been previously reported. The outcome is shown in Table~\ref{tbl:results2}.
We do not report results where running time is not mentioned. Some of the numbers in Table~\ref{tbl:results2} have higher accuracy than Falkon: this comes from the use of deep GPs which -- through a vast number of parameters -- can learn better data representations. 
Such models are intrinsically different in spirit from kernel methods, and we do not aim to compare with them specifically; they are reported in Table~\ref{tbl:results2} for the sake of completeness.

\section{Logistic Falkon Algorithm}\label{app:logistic}
In this section we provide some more details on how to derive fast algorithms with strong theoretical guarantees for smooth loss functions beyond squared loss. In particular, the main ideas from a theoretical and algorithmic viewpoint that we are going to recall here are developed in \cite{marteau2019beyond},~\cite{mareau19}. Our goal, as stated in the main text, is to make these ideas practical, by efficiently implementing and deploying the algorithms and making full use of the available computational architectures. In particular, we will focus on the following set of {\em generalized self concordant} loss functions:

\begin{algorithm}[t]
	\caption{Pseudocode for appr. Newton method with Falkon, for GSC losses (based on~\cite{mareau19}). \label{alg:GSC-falkon}}
	\begin{algorithmic}[1]
		\Function{GSC-Falkon}{$X \in \real{n\times d}, \bm{y} \in \real{n}, \lambda, m, t, T$}
		\State Set $\alpha_0 = 0 \in \real{m}$ and $\mu_0 > 0, q > 0$ according to~\cite{mareau19}.
		\State $X_m, \bm{y}_m \gets$ \Call{RandomSubsample}{$X,\bm{y}, m$}
		\For{$k \in \mathbb{N}$}
		\State $f_{k+1} \leftarrow$ \Call{WeightedFalkon}{$X,  \bm{y}, X_m, \bm{y}_m \mu_k, t, \alpha_{k}$}
		\State $\mu_{k+1} \leftarrow q\mu_{k}$
		\State Stop when $\mu_{k+1} < \lambda$ and set $\alpha_{last}\leftarrow \alpha_k$.
		\EndFor
		\Return $\widehat{\alpha} \leftarrow $ \Call{WeightedFalkon}{$X, \bm{y}, X_m, \bm{y}_m, \lambda, T, \alpha_{k}$}
		\EndFunction
	\end{algorithmic}
	
	$ $
	
	\begin{algorithmic}[1]
		\Function{WeightedFalkon}{$X \in \real{n\times d}, \bm{y} \in \real{n}, X_m \in \real{m \times d}, \bm{y}_m \in \real{m}, \lambda, t, \alpha_0 \in \real{m} $}
		\State $T, A \gets $ \Call{WeightedPreconditioner}{$X_m, \bm{y}_m, \alpha_0, \lambda$}
		\Function{LinOp}{$\bm{\beta} \in \real{m}$} 
		\State $\bm{v} \gets A^{-1}\bm{\beta}$
		\State $z \gets k(X,X_m) \bm{\beta}$ \Comment{predictions on the dataset}
		\State $D \gets \texttt{diag}[(\ell^{(2)}((z)_1, (\bm{y})_1), \dots, \ell^{(2)}((z)_n,(\bm{y})_n))]$
		\State $\bm{c} \gets k(X_m, X) D k(X, X_m)T^{-1}\bm{v}$
		\State \textbf{return} $A^{-\top} T^{-\top} \bm{c} + \lambda n \bm{v}$
		\EndFunction
		
		\State $R \gets A^{-\top}T^{-\top}k(X, X_m)\bm{y}$ 
		\State $\bm{\beta} \gets $ \Call{ConjugateGradient}{$\textsc{LinOp}, R,t, \alpha_0$} \Comment{CG solver starting from $\alpha_0$}
		\State \textbf{return} $T^{-1}A^{-1} \bm{\beta}$ 
		\EndFunction
	\end{algorithmic}
	
	$ $
	
	\begin{algorithmic}[1]
		\Function{WeightedPreconditioner}{$X_m \in \real{m \times d}, \bm{y}_m \in \real{m}, \alpha \in \real{m}, \lambda$}
		\State $K_{mm} \gets k(X_m, X_m)$ \Comment{Compute the kernel between inducing points} 
		\State $z \gets K_{mm} \alpha$ \Comment{predictions on the Nystr\"om points}
		\State $T \gets \chol(K_{mm})$ 
		\State $D \gets \texttt{diag}[(\ell^{(2)}((z)_1, (\bm{y}_m)_1), \dots, \ell^{(2)}((z)_m,(\bm{y}_m)_m))]$
		\State $K_{mm} \gets \nicefrac{1}{m} T D T^\top  + \lambda \eye$  
		\State $A \gets \chol(K_{mm})$ 
		\State \textbf{return} $T, A$ 
		\EndFunction
	\end{algorithmic}
	
	$ $
	
	Note: \textsc{LinOp} performs the multiplications via matrix-vector products.

\end{algorithm}%

\noindent{\bf Definition 1.} Generalized self-concordant (GSC) function~\cite{marteau2019beyond}
{\em
Let $\hilbert$ be a Hilbert space and let $z = (x,y)$ be an input-output pair. We say that $\ell_z: \hilbert \to {\mathbb R}$ is a generalized self-concordant function on ${\cal G} \subset \hilbert$, when ${\cal G}$ is a bounded subset of $\hilbert$ and $\ell_z$ is a convex and three times differentiable mapping on $\hilbert$ such that
\[ \textstyle \forall f \in \hilbert,~ \forall h,k \in \hilbert,~ \nabla^{(3)}\ell_{z}(f)[h,k,k] \leq \sup_{g \in {{\cal G}}}|g \cdot h| ~ \nabla^2 \ell_z(f)[k,k].\]
}%
Denote by $R$ the quantity $\sup_{g \in {\cal G}}\|g\| < \infty$. For many loss functions ${\cal G}$ is just the ball in ${\cal H}$ centered in zero and with radius $R > 0$, then $\sup_{g \in {\cal G}}|g \cdot h| = R\|h\|$).
The following loss functions, which are widely used in machine learning, are generalized self-concordant

\noindent{\bf Example 1.} (Application to finite-sum minimization~\cite{marteau2019beyond})
{\em
	The following loss functions are generalized self-concordant functions: 
	\\[.05cm]
	(a) Logistic  regression:
	$\ell_z(f)= \log(1 + \exp(  - y f(x)))$, where $z = (x,y)$ with $x \in X$ and $y \in \{-1,1\}$.
	\\[.05cm]
	(b) Robust regression: $\ell_z(f) = \varphi(f(x) - y)$ with $\varphi(u) = \log(e^u+e^{-u})$. Here $z = (x,y)$ with $x \in X$ and $y \in {\mathbb R}$
	\\[.05cm]
	(c) Softmax regression: $\ell_z(f) = \log(\sum_{j=1}^k [f(x)]_j) - [f(x)]_y$, where now $f:X \to \mathbb{R}^k$, $z = (x,y)$, with $y \in \{1,\dots,k\}$ and $v_j$ denotes the $j$-th column of $v \in \mathbb{R}^k$.
	\\[.05cm]
	(d) generalized linear models with bounded features, which include conditional random fields~(see more details in~\cite{marteau2019beyond}).
}
Note, in particular, that the loss functions above are generalized self concordant, but not {\em self concordant} as discussed in~\cite{marteau2019beyond}.

For the learning problem in Eq.~\eqref{eq:learning-problem} with generalized self-concordant loss functions, a strong theoretical result analogous to the one for kernel ridge regression \eqref{eq:bound} has been obtained~\cite{marteau2019beyond}. In particular, the regularized empirical risk minimization solution \eqref{eq:flambda_estimator} with generalized self-concordant losses achieves the bound
\begin{equation}\label{eq:bound-gsc} 
L(\hat{f}_\lambda) - \inf_{f \in \hilbert} L(f) = \O{n^{-1/2}},
\end{equation}
under standard regularity conditions on the learning problem and achieves fast learning rates similar to kernel ridge regression, considering more refined regularity conditions that are a natural extension of the conditions for kernel ridge regression~\cite{marteau2019beyond}.

The paper~\cite{mareau19} suggests to solve the regularized empirical risk minimization problem~\eqref{eq:flambda_estimator} for generalized self-concordant losses, by using a set of techniques that are extensions of the Falkon algorithm in~\cite{rudi2017falkon}. 
In particular, the problem is cast in terms of an approximate Newton method, with pseudocode shown in function \texttt{GSC-Falkon} of Algorithm~\ref{alg:GSC-falkon}. Nystr\"om method is used a first time to reduce the size of the problem, and then a second time to derive an approximate Newton step~\cite{mareau19}. Indeed a model of the form \eqref{eq:nystrom-solution} is considered and the preconditioner now plays the role of approximate Hessian, to perform the iterated approximation Newton. Given $(\tilde{x}_j, \tilde{y}_j)_{j=1}^m$ selected uniformly at random from the dataset, the approximate Hessian $\tilde{H}$ at the step $k$ is a weighted version of the Falkon preconditioner and has the form
$$ \tilde{H} = \frac{1}{m} T \tilde{D}_{k} T^\top + \mu_k I,$$
where $T$ is such that $T^\top T = K_{mm}$ (e.g. it is the Cholesky decomposition of $K_{mm}$) and $\tilde{D}_k \in \real{m\times m}$ is a diagonal matrix whose $j$th element is $\ell^{(2)}(f_k(\tilde{x}_j),\tilde{y}_j)$ where we assume that the loss function is $\ell(f(x),y)$ and the second derivative is taken with respect to the first variable.
As for Falkon, the approximate Hessian is never built explicitly, we compute instead its Cholesky decomposition in terms of the matrices $T, A$ as $\tilde{H}^{-1} = \tilde{P} \tilde{P}^\top$ with $\tilde{P} = T^{-1} A^{-1}$, see the function \texttt{WeightedPreconditioner} in Alg.~\ref{alg:GSC-falkon}. 
Then conjugate gradient is applied to the preconditioned problem, to solve the equation
\[
\tilde{P}^\top (K_{nm}^\top D_k K_{nm} + \lambda I) \tilde{P}\beta = \tilde{P}^\top K_{nm}^\top g_k. 
\]
where $D_k \in \real{n \times n}$ is a diagonal matrix whose $i$th element is $\ell^{(2)}(f_k(x_i),y_i)$ and $g_k \in \real{n}$ corresponds to $(g_k)_i = \ell^{(1)}(f_k(x_i), y_i)$.
To conclude, as proven in~\cite{mareau19}, to achieve the same learning rate of \eqref{eq:bound-gsc} and good practical performances,
\texttt{GSC-Falkon} (Alg.~\ref{alg:GSC-falkon}) needs to call \texttt{WeightedFalkon} only a small number of times with decreasing regularization parameters. Moreover, each time \texttt{WeightedFalkon} needs to execute only few iterations of the CG algorithm. 
The algorithm presented in Alg.~\ref{alg:GSC-falkon} has an important theoretical appeal as proved in~\cite{mareau19} since it is the fastest to date to achieve optimal learning rates for generalized self-concordant loss functions.
The goal of our work is to make it also appealing from a practical viewpoint. 
This requires efficiently implementing and deploying Alg.~\ref{alg:GSC-falkon}, making full use of the available computational architectures. Clearly the main bottlenecks here are the same of Falkon for squared loss and they are introduced and discussed in Section~\ref{sec:methods}.

\section{Out-Of-Core Algorithms}\label{app:ooc}
In this section we describe more in detail the out-of-GPU core algorithms for \begin{enumerate*}[label=\arabic*)]\item Cholesky decomposition of a positive definite matrix and \item multiplication of a triangular matrix by its transpose\end{enumerate*}. 
Both algorithms use a similar technique of dividing the input matrix in smaller tiles such that operations can be performed in-core on the individual tiles. 
Then the main challenges of such algorithms consist in choosing when to bring which tiles in-core, and how to do so in parallel, handling data-dependencies between different tiles.

We handle parallelism between multiple GPUs using a static work-allocation scheme where the input matrix is divided into block rows or columns (made up of several tiles), and each GPU is assigned one or more such rows (or columns) block-cyclically, to ensure that the workload is approximately balanced. 
Ensuring a balanced workload is tricky since the input matrices are triangular, and for example a row at the top of a lower-triangular matrix will have many more tiles than a row towards the bottom of said matrix. 
Smaller tile-sizes (so thinner block rows/columns) make each processor's workload more even, but -- in case the input matrix is not big enough -- they reduce overall GPU utilization.

\paragraph{Triangular matrix multiplication.}
\begin{algorithm}
	\caption{Out-of-core LAUUM operation on an upper-triangular matrix. The algorithm's inputs are matrix $U$, a synchronization object \texttt{barrier}, an array of arrays describing which row indices are assigned to which processor $\texttt{blockAllocs}$, and the number of tiles per side $N$. The function described below should be called for every available GPU with a different $\texttt{procId}$ value.}
	\label{alg:oocLauum}
	\begin{algorithmic}[1]
		\Function{OocLauum}{$U \in \real{n\times n}, \mathtt{barrier}, \mathtt{blockAllocs}, \mathtt{procId}, N$}
		\For{$i = 1, \dots, N$}
		\State $C\in \real{t\times t \cdot (N-i)} \gets \mathtt{ToGPU(procId,\ } \Big[U_{i, i}, \dots, U_{i, N} \Big]\mathtt{)}$
		\State \texttt{barrier.wait()}
		\For{$j \in \mathtt{blockAllocs[procId]}$}
		\If{$i = j$}
		\State $C_1 \gets C_1 C_1^\top$ \Comment{via \texttt{LAUUM}} \label{line:lauum}
		\If{$i \ne N$}
		\State $C_1 \gets C_1 +  C_{1:(N-i+1)} C_{1:(N-i+1)}^\top$ \Comment{via \texttt{SYRK}}
		\EndIf
		\ElsIf{$j > i$}
		\State $D \in \real{t\times t\cdot (N-j)} \gets \mathtt{ToGPU(procId,\ } \Big[ U_{j,j}, \dots, U_{j, N} \Big]\mathtt{)}$
		\State $C_{(j-i)} \gets C_{(j-i)}D_1^{\top}$ \Comment{via \texttt{TRMM}}
		\If{$j \ne N$}
		\State $C_{(j-i)} \gets C_{(j-i+1):(N-i+1)}D_{2:(N-j+1)}^\top$ \Comment{via \texttt{GEMM}}
		\EndIf
		\EndIf
		\State $U_{i,j} \gets \mathtt{FromGPU(procId,\ } C_{(j-i)}\mathtt{)}$ \label{line:write-back}
		\EndFor
		\EndFor
		\Return $U$
		\EndFunction
	\end{algorithmic}
\end{algorithm}

We begin by describing OOC triangular matrix multiplication, an operation which is known as LAUUM within the LAPACK library. Given an input upper triangular matrix $U\in\real{n\times n}$, we want to calculate the upper triangle of $UU^\top$ and store it in the upper part of $U$ (thus making this an in-place operation).
We divide $U$ in $N\times N$ tiles of size $t$ (uneven tile sizes are possible, and indeed necessary to support all input sizes, but omitted from the description for clarity), and we index all matrices by their tiles: $U_{2,2}$ is the square tile at the second block-row and second block-column of $U$.
The in-place LAUUM operation can be compactly described as $U_{i,j} = \sum_{k=j}^{N-1} U_{i,k}U_{j,k}^\top$ for $j \ge i$: to update a tile of $U$ we need to multiply two block-rows of the original matrix. However, we can exploit the triangular structure of some of the above matrix multiplications to improve performance: for example, when $i = j$ it is possible to split the update into two parts $U_{i,i} = U_{i,i}U_{i,i}^\top + \sum_{k=i}^N U_{i,k}U_{i,k}^\top$ where the first part consists of an in-core LAUUM operation and the second of a symmetric matrix multiplication (BLAS routine SYRK) which can be up to twice as fast as the general matrix multiplication routine. 
Similarly, for $i < j$, $U_{i,j} = U_{i,j}U_{j,j}^\top + \sum_{k=j + 1}^N U_{i,k}U_{j,k}^\top$ where the first part can use the TRMM routine from the BLAS library and the second must use the generic GEMM routine.
To avoid overwriting parts of $U$ which are still needed for the updates -- especially in a multi-GPU setting -- the rows of $U$ are to be updated one at a time, from top to bottom. To ensure synchronization between multiple GPUs we use a thread barrier so that all GPUs start updating a given row after having loaded its original, non-updated version in GPU memory.
GPU memory requirements for Algorithm~\ref{alg:oocLauum} are two block-columns (i.e. $2Nt^2$ numbers). 
As discussed above, rows are assigned to GPUs in a 1D block-cyclic way. Such allocations are recorded in the \texttt{blockAllocs} variable.

An adaptation of Algorithm~\ref{alg:oocLauum} is possible when in-place operation is not needed: it is sufficient to remove the synchronization barrier, and change line~\ref{line:write-back} to write the output to a different matrix.

\paragraph{Cholesky decomposition.}
\begin{algorithm}
	\caption{Out-of-core, in-place Cholesky decomposition of symmetric positive definite matrix $A$. The lower triangle of $A$ will be overwritten by $L$ such that $L^\top L = A$. The function \texttt{OocPotrf} should be called for each available GPU with different values of the \texttt{procId} variable to parallelize the decomposition across GPUs. The inputs are the same as for Algorithm~\ref{alg:oocLauum} but for work-table $T\in\mathbb{Z}^{N\times N}$ whose values are atomically updated by the different GPU processes to ensure synchronization.}
	\label{alg:oocPotrf}
	\begin{minipage}[t]{0.62\linewidth}{
			\begin{algorithmic}[1]
				\Function{OocPotrf}{$A\mathtt{, blockAllocs, procId,\ }T\mathtt{,\ }N$}
				\For{$i = 1, \dots, N$}
				\If{$i \in \mathtt{blockAllocs[procId]}$}
				\State $B \gets \mathtt{Load(}A\mathtt{,\ }T\mathtt{,\ }i\mathtt{,\ }j\mathtt{,\ }i\mathtt{)}$
				\State $B \gets \mathtt{POTRF(}B\mathtt{)}$
				\State $A_{i,i} \gets \mathtt{Write(}B\mathtt{,\ }T\mathtt{,\ }i\mathtt{,\ }i\mathtt{)}$
				\EndIf
				
				\For{$j \in \mathtt{blockAllocs[procId]}$}
				\If{$j \le i$}
				\State $\mathtt{continue}$
				\EndIf
				\State $B \gets \mathtt{Load(}A\mathtt{,\ }T\mathtt{,\ }i\mathtt{,\ }i\mathtt{,\ }i+1\mathtt{)}$
				\State $C \gets \mathtt{Load(}A\mathtt{,\ }T\mathtt{,\ }j\mathtt{,\ }i\mathtt{,\ }i\mathtt{)}$
				\State $C \gets C(B^{-1})^\top$ \Comment{via \texttt{TRSM}}
				\State $A_{j, i} \gets \mathtt{Write(}C\mathtt{,\ }T\mathtt{,\ }j\mathtt{,\ }i\mathtt{)}$
				\EndFor
				
				\For{$j \in \mathtt{blockAllocs[procId]}$}
				\If{$j \le i + 1$}
				\State $\mathtt{continue}$
				\EndIf
				\State $C \gets \mathtt{Load(}A\mathtt{,\ }T\mathtt{,\ }j\mathtt{,\ }i\mathtt{,\ }i+1\mathtt{)}$
				\For{$y = i, \dots j$}
				\State $E \gets \mathtt{Load(}A\mathtt{,\ }T\mathtt{,\ }j\mathtt{,\ }y\mathtt{,\ }i\mathtt{)}$
				\If{$y = j$}
				\State $E \gets E - CC^\top$ \Comment{via \texttt{SYRK}}
				\Else
				\State $D \gets \mathtt{Load(}A\mathtt{,\ }T\mathtt{,\ }y\mathtt{,\ }i\mathtt{,\ }i+1\mathtt{)}$
				\State $E \gets E - DC^\top$ \Comment{via \texttt{GEMM}}
				\EndIf
				\State $A_{j, y} \gets \mathtt{Write(}E\mathtt{,\ }T\mathtt{,\ }j\mathtt{,\ }y\mathtt{)}$
				\EndFor
				\EndFor
				\EndFor
				\EndFunction
			\end{algorithmic}%
	}\end{minipage}%
	\adjustbox{valign=t}{
		\begin{minipage}[t]{0.37\linewidth}{
				\begin{algorithmic}[1]
					\setcounter{ALG@line}{34}
					\Function{Load}{$A, T, i, j, \mathtt{exp}$}
					\While{$T_{i,j} < \mathtt{exp}$}
					\State \texttt{wait}
					\EndWhile
					\State \Return $\mathtt{ToGPU(}A_{i,j}\mathtt{)}$
					\EndFunction
					\item[]  
					\Function{Write}{$G, T, i, j$}
					\State $T_{i,j} \gets T_{i,j} + 1$
					\State \Return $\mathtt{FromGPU(}G\mathtt{)}$
					\EndFunction
				\end{algorithmic}%
				\vfill%
	}\end{minipage}}%
\end{algorithm}

We want to decompose positive definite matrix $A$ into lower triangular matrix $L$ such that $L^\top L = A$. But $A$ does not fit entirely in GPU memory, and potentially more than one GPU is available.
As before it is convenient to subdivide $A$ into smaller tiles such that the tiles fit in GPU memory. 
\begin{equation*}
\begin{pmatrix}
A_{1,1} &         &         \\
A_{2,1} & A_{2,2} &         \\
\vdots  & \ddots  &         \\
A_{n,1} & \dots   & A_{n,n}
\end{pmatrix} 
=
\begin{pmatrix}
L_{1,1} &         &         \\
L_{2,1} & L_{2,2} &         \\
\vdots  & \ddots  &         \\
L_{n,1} & \dots   & L_{n,n}
\end{pmatrix}
\begin{pmatrix}
L_{1,1}^T & L_{2,1}^T & \dots & L_{n,1}^T \\
& L_{2,2}^T & \dots & L_{n,2}^T \\
&           & \ddots& \vdots    \\
&           &       & L_{n,n}^T
\end{pmatrix}
\end{equation*}
Then the in-place decomposition may proceed column-wise across matrix $A$, where each column update requires three steps.
The first step is to use the in-core \texttt{POTRF} function from cuSOLVER~\cite{cusolver} on a single tile. Then, a triangular solution step is used to update the remaining rows of the first column (taking the first column as an example $A_{j,1} = L_{j,1}L_{1,1}^\top, 1 < j < N$, so clearly $L_{j,1} = A_{j,1}(L_{1,1}^{-1})^\top$). This can be done by using the \texttt{TRSM} operation from any GPU BLAS implementation.
Finally, the \textit{trailing} submatrix must be updated with those terms which can be computed from the current column, so that after this last step such column is not needed anymore. This step consists of running $A_{ij} = A_{ij} - L_{i,1}L_{j,1}^\top$ where if $c$ is the current column $i > c,\quad c < j \le i$ (refer to Figure~\ref{fig:potrf} for a more intuitive picture).

Running this algorithm in parallel requires dealing with several data dependencies in-between tiles, and in general it will not be possible to achieve perfect parallelism due to the inherently serial step of performing the Cholesky decomposition of the first tile in a column. 
We avoid coarse synchronization mechanisms such as the thread barrier which was used for the LAUUM OOC implementation, since we found they could introduce very high waiting times (barriers would be needed after each of the three steps of the algorithm to ensure proper synchronization). 
Our solution, which somewhat follows~\cite{ltaief10magma}, uses an integer table $T$ with one entry per tile, which is shared between all GPU threads. The entries of $T$ represent the current state of each tile: basically how many times the tile has been updated. 
Since we use a static row-cyclic work allocation like for the triangular matrix multiplication, each thread knows the expected state of a tile for each step (e.g. to perform the first step on tile $A_{c,c}$ the tile must have been updated exactly $c$ times). So it can wait until such state has been reached, then read the tile into GPU memory, perform the update, write back the tile to main RAM, and increment the corresponding entry in $T$. Such a scheme is implemented in Algorithm~\ref{alg:oocPotrf} with the help of the \texttt{Load} and \texttt{Write} sub-routines.
Further optimizations are possible by being careful about which tiles are swapped in and out of GPU memory and at what times, overlapping computation with memory transfers when possible.
Such optimizations generally require to increase the total memory allocated on the GPU, thus decreasing the maximum possible tile-size.

\end{document}